\newif\ifsingle

\newif\ifFullVersion
\FullVersiontrue 

\ifsingle
\documentclass[11pt,draftclsnofoot, onecolumn]{IEEEtran}		
\else		
\documentclass[10pt,final, twocolumn]{IEEEtran}
\fi

\usepackage{times}
\usepackage{amsmath,dsfont}
\usepackage{amssymb,amsthm}
\usepackage{epsfig,verbatim} 
\usepackage{setspace}
\usepackage{color}
\usepackage{cite}
\usepackage{epstopdf}
\usepackage{graphics}
\usepackage{accents}
\usepackage{acronym}
\usepackage[bookmarks,colorlinks]{hyperref}
\usepackage{booktabs}
\usepackage{mathtools} 
\usepackage{enumitem}
\usepackage{wrapfig}
\usepackage{subcaption}

\usepackage[ruled,linesnumbered]{algorithm2e}  
\SetKwInput{KwData}{\textbf{Init}}

\definecolor{NewColor}{rgb}{0,0,0} 

\newcommand{\myVec}[1]{{\boldsymbol{#1}}}
\newcommand{\myMat}[1]{{\boldsymbol{#1}}}
\newcommand{\mySet}[1]{\mathcal{#1}}
\newcommand{\myX}{{\myVec{x}}}			 		

\newcommand{\FwdMsg}[2]{\overrightarrow{\mu}_{\! #2}}
\newcommand{\BwdMsg}[2]{\overleftarrow{\mu}_{\! #2}}

\newcommand{\Pdf}[1]{P_{ { #1}} }

\newcommand{\Mem}{l}			 			
\newcommand{\Blklen}{t}			 			
\newcommand{\Blkset}{\mySet{T}}
	
\newcommand{\NonStatBlk}{t_b}

\acrodef{adc}[ADC]{analog-to-digital convertor}
\acrodef{cs}[CS]{compressed sensing}
\acrodef{bp}[BP]{belief propagation}
\acrodef{sp}[SP]{sum-product}
\acrodef{dtft}[DTFT]{discrete-time Fourier transform}
\acrodef{dnn}[DNN]{deep neural network} 
\acrodef{csi}[CSI]{channel state information}
\acrodef{map}[MAP]{maximum a-posteriori probability}
\acrodef{snr}[SNR]{signal-to-noise ratio}
\acrodef{bs}[BS]{base station} 
\acrodef{em}[EM]{expectation maximization} 
\acrodef{iot}[IOT]{Interent of Things}
\acrodef{mimo}[MIMO]{multiple-input multiple-output}
\acrodef{mse}[MSE]{mean-squared error}
\acrodef{pdf}[PDF]{probability density function}
\acrodef{rv}[RV]{random variable}
\acrodef{fec}[FEC]{forward error correction} 
\acrodef{lti}[LTI]{linear time-invariant}
\acrodef{wss}[WSS]{wide-sense stationary}
\acrodef{psd}[PSD]{power spectral density}
\acrodef{ser}[SER]{symbol error rate} 
\acrodef{ber}[BER]{bit error rate} 
\acrodef{sgd}[SGD]{stochastic gradient descent} 
\acrodef{isi}[ISI]{intersymbol interference}  
\acrodef{awgn}[AWGN]{additive white Gaussian noise} 
\acrodef{ut}[UT]{user terminal}  
\acrodef{ml}[ML]{machine learning}  
\acrodef{rnn}[RNN]{recurrent neural network}
\acrodef{sbrnn}[SBRNN]{sliding bidirectional \ac{rnn}}
\acrodef{bpsk}[BPSK]{binary phase shift keying}
\acrodef{rs}[RS]{Reed-Solomon}
\acrodef{crc}[CRC]{cyclic redundency check}
\acrodef{fc}[FC]{fully-connected}
\acrodef{hmm}[HMM]{hidden Markov model}
\acrodef{loo}[LOO]{leave-one-out}

\ifsingle
\newcommand{\figWidth}{0.75\columnwidth}

\newcommand{\includefig}[1]{\includegraphics[width = 0.75\columnwidth]{#1} 	\vspace{-0.2cm}}
\else
\newcommand{\figWidth}{\columnwidth}

\newcommand{\includefig}[1]{\includegraphics[width = \columnwidth]{#1} 	\vspace{-0.2cm}}

\fi 

\IEEEoverridecommandlockouts

\title{Learned Factor Graphs for Inference \\from Stationary Time Sequences
}
\author{
	\IEEEauthorblockN{Nir Shlezinger, Nariman Farsad, Yonina C. Eldar, and A. J. Goldsmith\\
	} 
	\thanks{
		    Parts of this work were presented in the IEEE International Symposium on Information Theory (ISIT) 2020 \cite{shlezinger2020data}. 
	This work was supported in part by the Israeli Innovation Authority through the 5G-WIN consortium, the Benoziyo Endowment Fund for the Advancement of Science, QuantERA grant  C’MON-QSENS!, the European Union’s Horizon 2020 research and innovation program under grant No. 646804-ERC-COG-BNYQ,  the Israel Science Foundation under grant No. 0100101, and by the Office of the Naval Research under grant No. 18-1-2191. 
		N. Shlezinger is with the School of ECE, Ben-Gurion University of the Negev, Beer-Sheva, Israel (e-mail: nirshl@bgu.ac.il).
		N. Farsad is with Department of CS, Ryerson University, Toronto, Canada (e-mail: nfarsad@ryerson.ca). 
		Y. C. Eldar is with the Faculty of Math and CS, Weizmann Institute of Science, Rehovot, Israel (e-mail: yonina.eldar@weizmann.ac.il). 	
		A. J. Goldsmith is with the Department of EE, Princeton University, Princeton, NJ (e-mail: goldsmith@princeton.edu).  }

	\vspace{-1.0cm}
	
}
\vspace{-0.75cm}

\begin{document}
	
	\maketitle
	\pagestyle{plain}
	\thispagestyle{plain}
	
	\begin{abstract}
	The design of methods for inference from time sequences has traditionally relied on statistical models that describe the relation between a latent desired sequence and the observed one. A broad family of model-based algorithms have been derived to carry out  inference at controllable complexity using recursive computations over the factor graph representing the underlying distribution. An alternative model-agnostic approach utilizes machine learning (ML) methods. Here we propose a framework that combines model-based  algorithms and data-driven ML tools for stationary time sequences. In the proposed approach, neural networks are developed to separately learn specific components of a factor graph describing the distribution of the time sequence, rather than the complete inference task. By exploiting stationary properties of this distribution, the resulting approach can be applied to sequences of varying temporal duration. Learned factor graphs can be realized using compact neural networks that are trainable using small training sets, or alternatively,  be used to improve upon existing deep inference systems. We present an inference algorithm based on learned stationary factor graphs,  which learns to implement the sum-product scheme from labeled data, and can be applied to sequences of different lengths. Our experimental results demonstrate the ability of the proposed learned factor graphs to learn from small training sets to carry out accurate inference for sleep stage detection using the Sleep-EDF dataset, as well as for symbol detection in digital communications with unknown channels. 
	\end{abstract}
	
\vspace{-0.4cm}
\section{Introduction}
\vspace{-0.1cm} 
A multitude of practical problems involve inference from time sequences. The need to accurately estimate a hidden time series from a measured signal is frequently encountered in signal processing, communications, control, finance, and various other fields.  Traditional algorithms, such as those based on the \ac{map} rule, are  model-based, namely, they carry out inference based on complete knowledge of the underlying statistical model relating the desired time series and the observed one.  The joint distribution of a large family of time sequences encountered in practice can be factorized, which facilitates inference at reduced complexity by representing their distribution as a Forney-style factor graph \cite{forney2001codes,loeliger2004introduction}, referred to henceforth as a {\em factor graph} for brevity. {In factor graphs, variables are represented as edges connected to function nodes forming a graphical representation of joint distribution measures. When this graph is cycle-free, it can be used to evaluate marginal distributions in an efficient manner, i.e., with complexity that only grows linearly with the number of variables.}  As a result, many important model-based algorithms, such as the Viterbi algorithm \cite{viterbi1967error}, the \ac{sp} method \cite{kschischang2001factor} also known as \ac{bp} \cite{pearl1986fusion}, the BCJR detector \cite{bahl1974optimal}, {the Baum-Welch scheme for estimating the parameters of \acp{hmm} \cite{baum1970maximization},} and the Kalman filter \cite[Ch. 7]{haykin2005adaptive}, all process time sequences via recursive computations over a factor graph 
\cite{loeliger2007factor}.

Often in practice, the underlying statistical model relating the observations and the desired time series is highly complex or poorly understood. In such cases, model-based algorithms, which are typically sensitive to inaccurate knowledge of the underlying statistics, cannot be reliably applied, and model-agnostic data-driven schemes are preferable. Consequently, recent years have witnessed extensive interest in the application of \ac{ml}, and particularly of \acp{dnn}, for time sequence inference, with various architectures proposed to exploit the presence of  temporal correlation \cite{gamboa2017deep,fawaz2019deep,vaswani2017attention}. However, training these deep architectures typically requires a massive amount of labeled data, which may not always be available. Furthermore, applying inference using highly parameterized \acp{dnn} may not be feasible on  devices with limited hardware capabilities. 

The individual challenges of model-based signal processing and model-agnostic \ac{ml} has given rise to various hybrid systems  combining \ac{ml} and model-based algorithms \cite{shlezinger2020model}. Such hybrid model-based/data-driven attempt to benefit from the best of both worlds. These include the usage of \acp{dnn} to learn a possibly analytically intractable regularization in compressed sensing applications \cite{bora2017compressed,van2018compressed}, as well as the use of deep denoisers in regularized optimization via plug-and-play networks \cite{venkatakrishnan2013plug,ahmad2020plug}. A systematic strategy to combine model-based algorithms and \ac{ml} is based on deep unfolding or unrolling \cite{gregor2010learning, hershey2014deep, monga2019algorithm, balatsoukas2019deep}. Deep unfolding sets the layers of a \ac{dnn} in light of the iterations of some iterative optimization algorithm, while using the resulting unfolded network for the complete inference task.  Deep unfolding typically requires knowledge of the underlying model, up to perhaps some unknown parameters, in order to unfold the optimization method. Unfolding commonly results in a highly parameterized \ac{dnn} utilized for the complete inference task, whose architecture is inspired by a model-based algorithm. 

In this work, we propose an alternative strategy that combines model-based signal processing algorithms based on factor graph computations with data-driven \ac{ml} tools. Here, instead of using \acp{dnn} for inference, they are utilized to {\em learn only the function nodes of the factor graph}, which in turn is used for inference via conventional factor graph methods, such as the \ac{sp} algorithm. This approach builds upon the fact that the statistical behavior of time sequences can often be approximated using stationary factorizable distributions, which allows incorporating  domain knowledge in the structure of the graph while learning its nodes from data.  
This results in a hybrid model-based/data-driven inference scheme that only requires prior knowledge of the factorization of the underlying distribution. This is in contrast to deep unfolding where  the distribution in parametric form is typically required. Moreover, the hybrid scheme can also incorporate additional domain knowledge in its learned nodes.  As opposed to previous works that used highly-parameterized deep architectures to represent messages along a factor graph, and trained the overall system in an end-to-end manner \cite{bruna2013spectral,zheng2015conditional,lin2015deeply,gilmer2017neural, zhang2019factor, yoon2019inference, satorras2019combining, satorras2020neural}, our strategy learns the function nodes separately from the task, exploiting stationarity by reusing a \ac{dnn} for multiple function nodes. Consequently, this approach can use relatively compact networks that are trained with  small training sets and employed on hardware-limited devices, as well as to improve upon existing \ac{dnn} architectures by utilizing them for learning the factor graph instead of for inference. 
Furthermore,  the same learned factor graph may be used for sequences of varying length, as well as combined with  multiple inference algorithms.

In particular, we present a data-driven inference scheme based on learned factor graphs that learns to implement the \ac{sp} method over factor graphs of stationary Markovian time sequences  from labeled data. While the  \ac{sp} scheme requires accurate knowledge of the underlying statistical model, its data-driven implementation allows this algorithm to be utilized in scenarios involving time sequences with complex and possibly analytically intractable distributions.   We detail how such hybrid model-based/data-driven inference is derived from the \ac{sp} algorithm by learning the underlying factor graph. We also show  that, by assuming  stationarity,  the complete factor graph can be learned using a single relatively compact neural network. 
We discuss how ViterbiNet, proposed in \cite{shlezinger2019viterbinet} for real-time data-driven symbol detection in digital communications, can be obtained as a special case of our framework, and in fact be implemented using the same learned factor graph as that used for \ac{sp}-based inference. We then discuss how the ability to learn factor graphs of stationary distributions from small training sets may be exploited to facilitate adaptation to blockwise statistical variations when some future indication on the inference correctness is available, as in, e.g., coded communications setups.

We evaluate the usage of learned factor graphs for sleep pattern prediction as well as in a digital communications setup with unknown channel settings.  For sleep pattern prediction, we use the Sleep-EDF dataset \cite{kemp2013sleep}, and show how using a neural predictor as a learned node in the \ac{sp} method  improves  the accuracy by $4\%$ compared with using it for inference. We also demonstrate how this facilitates the usage of compact networks trainable with small data sets, enabling training using only part of the data of a single patient, while achieving accurate inference on its remaining data.
For the communications setup, we demonstrate that inference over learned factor graphs is capable of approaching the performance of the \ac{map} detector, which requires full knowledge of the underlying statistical model, while achieving improved robustness to model uncertainty compared with the conventional \ac{sp} algorithm. 
Furthermore, we demonstrate that by utilizing compact networks that are trainable with small training sets, learned factor graphs can be tuned to accurately track temporal variations in the statistical model via online training.

The rest of this paper is organized as follows:  In Section~\ref{sec:Model}, we detail the problem of inference over stationary time sequences, and briefly review model-based factor graph methods. Section~\ref{sec:Inference} details the proposed framework for inference over learned factor graphs by deriving it from the \ac{sp} algorithm applied to Markovian signals. Experimental results are presented in Section~\ref{sec:sims}. Finally, Section~\ref{sec:Conclusions} provides concluding remarks.

 Throughout the paper, we use upper-case letters for \acp{rv}, e.g. $X$.
Boldface lower-case letters denote vectors, e.g., ${\myVec{x}}$ is a deterministic vector, and $\myVec{X}$ is a random vector; 
the $i$th element of ${\myVec{x}}$ is written as $x_i$. 
The probability measure of an \ac{rv} $X$ evaluated at $x$ is denoted $P_X(x)$, while $\mathcal{N}(\cdot,\cdot)$ represents the Gaussian distribution. We use caligraphic letters for sets, e.g., $\mySet{X}$, where $|\mySet{X}|$ is the cardinality of a finite set $\mySet{X}$, while 
$\mathbb{R}$ denotes the set of real numbers. Finally, for a sequence $\{x_i\}$ and integers $ i_1 < i_2$ we use $\myVec{x}_{i_1}^{i_2}$ to denote the stacking $[x_{i_1},x_{i_1+1}, \ldots, x_{i_2}]^T$ while $\myVec{x}^{i_2}\triangleq \myVec{x}_{1}^{i_2}$. 
\vspace{-0.2cm}
\section{System Model}
\label{sec:Model}
\vspace{-0.1cm}
\ifFullVersion
In this section, we present the system model for which we propose the concept of data-driven factor graphs in Section \ref{sec:Inference}. 
We begin by formulating the considered time series inference problem in Subsection \ref{subsec:FGModSequences}. We then discuss the model-based approach for the problem at hand in Subsection \ref{subsec:Model Inference}, after which we briefly review conventional model-based factor graph methods in Subsection \ref{subsec:FGModel}.
\fi	
\vspace{-0.2cm}
\subsection{Problem Formulation}
\label{subsec:FGModSequences}
\vspace{-0.1cm}	 
We consider the problem of recovering a desired time series $\{S_i\}$ taking values in a  set $\mySet{S}$ from an observed sequence $\{Y_i\}$ taking values in a set $\mySet{Y}$. The subscript $i$ denotes the time index. The joint distribution of $\{S_i\}$ and $\{Y_i\}$ obeys an $\Mem$th-order Markovian stationary model,
\begin{align}
&\Pdf{Y_i, S_i | \{Y_j, S_j\}_{j < i}}\left(y_i, s_i | \{y_j, s_j\}_{j < i}\right) \notag \\
& \qquad =  \Pdf{Y_i | \myVec{S}_{i-\Mem}^{i}}\left(y_i|\myVec{s}_{i-\Mem}^{i}\right) \Pdf{S_i |  \myVec{S}_{i-\Mem}^{i-1}}\left(s_i | \myVec{s}_{i-\Mem}^{i-1}\right),
\label{eqn:FiniteMemory}
\end{align}
	for some integer $\Mem \geq 1$, representing the memory of the sequences. 
Consequently, when the initial state $\myVec{S}_{-\Mem}^{0}$ is given,  the joint distribution of  $\myVec{Y}^\Blklen$ and $\myVec{S}^\Blklen$  satisfies 
\begin{equation}
\label{eqn:MarkovModel}
\Pdf{\myVec{Y}^\Blklen,\myVec{S}^\Blklen}(\myVec{y},\myVec{s}) \!=\! 
\prod_{i=1}^{\Blklen} \Pdf{Y_i | \myVec{S}_{i-\Mem}^{i}}\!\left(y_i|\myVec{s}_{i-\Mem}^{i}\right)\! \Pdf{S_i |  \myVec{S}_{i-\Mem}^{i-1}}\!\left(s_i | \myVec{s}_{i-\Mem}^{i-1}\right),
\end{equation}
for any fixed sequence length $\Blklen >0$. 
{The joint distribution in \eqref{eqn:MarkovModel} is a special case of an $\Mem$th order Markov model, where given $\myVec{S}_{i-\Mem}^{i}$, the observed sequence $Y_i$ does not depend on the past observations $\myVec{Y}_{i-\Mem}^{i-1}$.}
{In principle, the $\Mem$th order Markov model in  \eqref{eqn:MarkovModel} can be expressed as an order-one Markov model by replacing $S_i$ with the multivariate $ \{\myVec{S}_{i-\Mem+1}^{i}\}_{i=1}^{\Blklen}$. However, since we focus on the recovery of the sample $S_i$ from $\myVec{Y}^\Blklen$ (rather than the stacking of $\Mem$ samples $\myVec{S}_{i-\Mem+1}^{i}$), we keep the above formulation without limiting our attention to the case where $\Mem=1$.}

%
	In general, the above statistical relationship can change over time. Here, we assume that the marginal distribution of the desired sequence $\{S_i\}$ remains static over time, i.e., $P_{S_i|\myVec{S}_{i-\Mem}^{i-1}}(\cdot)$  does not depend on $i$. We allow the conditional \ac{pdf} in \eqref{eqn:MarkovModel} to evolve over time in the following manners:
	\begin{itemize}
		\item {\em Stationary sequence} - the conditional \ac{pdf} in \eqref{eqn:MarkovModel} remains invariant over time, i.e.,  $\Pdf{Y_i|S_i}(\cdot)$ does not depend on the time index $i$.
		\item {\em Blockwise stationary} - the conditional \ac{pdf} $\Pdf{Y_i|S_i}(\cdot)$ changes every $\NonStatBlk$ time instances. 
	\end{itemize}  
%
The stationarity assumption implies that within a given block of  $\NonStatBlk$ time instances, \eqref{eqn:MarkovModel} represents the joint distribution as the product of the same function with different arguments. 

The statistical relationship in \eqref{eqn:MarkovModel} accurately represents a broad range of problems encountered in practice, including inference from bio-medical signals \cite{vullings2010adaptive,jiang2019robust} as well as symbol detection in digital communications \cite[Ch. 3]{goldsmith2005wireless}. The common aspect of these problems is the presence of temporal correlation, implying that information regarding a state variable $S_i$ is contained not only in its corresponding observation  $Y_i$, but also in its preceding and subsequent measurements.  

Our goal is to design a system that learns to reliably infer a block of desired variables $\myVec{S}^\Blklen$ from its corresponding observations  $\myVec{Y}^\Blklen$ for arbitrary blocklength $\Blklen$. The system learns its inference mapping using a {data set comprised of a sequence of $n_t$ labeled samples denoted $\{s_k, y_k\}_{k=1}^{n_t}$}, as well as prior knowledge that the joint distribution obeys the factorization in \eqref{eqn:MarkovModel}. Nonetheless, it is emphasized that distribution functions in \eqref{eqn:MarkovModel}, e.g., $\Pdf{Y_i | \myVec{S}_{i-\Mem}^{i}}$, are unknown and may not be analytically tractable. The presence of data as well as partial domain knowledge motivates the design of hybrid model-based/data-driven schemes.

\vspace{-0.2cm}
\subsection{Model-Based Inference}
\label{subsec:Model Inference}
\vspace{-0.1cm}
When the joint distribution of $\myVec{S}^\Blklen$ and $\myVec{Y}^\Blklen$ is a-priori known and can be computed, the  inference rule that minimizes the symbol error probability for each time instance is the \ac{map} detector,
\begin{align}
\hat{s}_i\left( \myVec{y}^\Blklen\right)  &= \mathop{\arg \max}\limits_{s \in \mySet{S}}\Pdf{S_i|\myVec{Y}^\Blklen}(s|\myVec{y}^\Blklen),
\label{eqn:MAP0}
\end{align}
for each $i\in\{1,\ldots,\Blklen\} \triangleq  \Blkset$.
This rule can be efficiently evaluated for finite-memory distributions using the \ac{sp} algorithm \cite{kschischang2001factor}.    	
An alternative common inference rule 
is the maximum likelihood sequence detector, given by  
\begin{align}
\hat{\myVec{s}}^{ \Blklen}\left( \myVec{y}^{ \Blklen}\right)  
&\triangleq \mathop{\arg \max}_{\myVec{s}^{ \Blklen} \in \mySet{S}^\Blklen } \Pdf{\myVec{Y}^{ \Blklen} | \myVec{S}^{ \Blklen}}\left( {\myVec{y}^{ \Blklen} | \myVec{s}^{ \Blklen}}\right).
\label{eqn:ML0} 
\end{align}
{Unlike \eqref{eqn:MAP0}, the maximum likelihood sequence detector \eqref{eqn:ML0} has no optimality guarantee in general. However, similarly to \eqref{eqn:MAP0}, the inference rule in \eqref{eqn:ML0} is amenable to efficient iterative computation. In particular,} 
for sequences obeying the structure \eqref{eqn:MarkovModel}, the detector \eqref{eqn:ML0} can be computed efficiently using the Viterbi algorithm \cite{viterbi1967error,forney1973viterbi}. Viterbi detection allows real-time inference, i.e., it operates in a sequential manner and uses the partial vector $\myVec{y}^{i+\Mem -1}$, instead of the complete observations $ \myVec{y}^{ \Blklen}$, when recovering $S_i$.

Both the \ac{sp} method and the Viterbi scheme are model-based algorithms that employ 
recursive computations over the underlying factor graph encapsulating the joint distribution of $\myVec{S}^\Blklen$ and $\myVec{Y}^\Blklen$. Consequently, to design a system capable of learning from data how to carry out such inference, we first review factor graph methods in the following subsection.

\vspace{-0.2cm}
\subsection{Factor Graphs Methods}
\label{subsec:FGModel}
\vspace{-0.1cm} 
In the following we provide a brief introduction to factor graphs. We then review the \ac{sp} algorithm for computing marginal distributions as a representative method for efficient inference over factor graphs \cite{kschischang2001factor}.    

A Forney-style factor graph, referred to henceforth as factor graph, is a graphical representation of the factorization of a function of several variables \cite{loeliger2007factor}, commonly a joint distribution measure. Its main advantage over alternative graphical models of joint distributions, such as Bayesian (belief) networks \cite{pearl2014probabilistic} and junction graphs \cite{aji2000generalized}, stems from its suitability to hierarchical models and its resulting simple formulation of the \ac{sp} message passing algorithm \cite{loeliger2004introduction}. To present the concept of factor graphs and their usage, consider a $\Blklen \times 1$ random vector $\myVec{X} \in \mySet{X}^\Blklen$ where $\mySet{X}$ is a finite set, i.e., the entries of $\myVec{X}$, denoted $\{X_i\}$, are discrete \acp{rv}. The joint distribution of $\myVec{X}$, $P_{\myVec{X}}(\myX)$, is factorizable if it can be represented as the product of $m$ functions $\{f_k(\cdot)\}_{k=1}^{m}$, i.e., there exist some partition variables $\{\mySet{V}_k\}_{k=1}^{m}$, $\mySet{V}_k \subset \{x_1, \ldots, x_\Blklen\}$, {which are not subsets of one another,} such that 
	\begin{equation}
P_{\myVec{X}}(\myX) = \prod_{k=1}^{m} f_k(\mySet{V}_k).
	\label{eqn:Decomp}
	\end{equation} 
	In order to represent \eqref{eqn:Decomp} as a factor graph, the functions $\{f_k(\cdot)\}_{k=1}^{m}$ should be set such that each variable $x_i$ appears in no more than two partitions\footnote{A factorization in which a variable appears in more than two factors can always be modified to meet the above constraint by introducing additional variables and identity factors, see \cite{loeliger2004introduction}.}  $\{\mySet{V}_k\}_{k=1}^{m}$. Subject to this assumption, 
the  distribution $P_{\myVec{X}}(\myX)$ can be described as a factor graph with $m$  nodes, which are the functions  $\{f_k(\cdot)\}_{k=1}^{m}$, while the variables $\{x_i\}_{i=1}^{\Blklen}$ represent edges or half-edges. {In the sequel we focus on partitions in which the resulting graphical representation is cycle-free.} 

A major motivation for representing joint distributions via factor graphs is that they allow some  statistical computations to be carried out with reduced complexity. One of the most common methods to exploit factorization via factor graphs for reduced complexity inference is the \ac{sp} algorithm, that evaluates a marginal distribution from a factor graph representation of a joint probability measure \cite{kschischang2001factor}. To formulate the \ac{sp} method, consider {for simplicity} a factorized distribution in which the ordering of the partitions $\{\mySet{V}_i\}$ corresponds to the order of the variables $\{x_i\}$, e.g., $\mySet{V}_1 = \{x_1, x_2\}$, $\mySet{V}_2 = \{x_2, x_3, x_4\}$, $\mySet{V}_3 = \{x_4, x_5\}$, etc. Furthermore, {as noted above,} assume that the factor graph does not contain cycles.\footnote{\label{ftn:cycles} In the presence of cycles in the graph, the \ac{sp} algorithm does not compute the \ac{map} rule, but can approximate it iteratively  \cite{weiss2001optimality}. Here, we focus  on the standard application for cycle-free graphs \cite{loeliger2004introduction,loeliger2007factor}.} In this case, the marginal distribution of an \ac{rv} $X_i$ whose corresponding variable appears in the partitions $\mySet{V}_{j}$ and $\mySet{V}_{j+1}$ can be computed from the joint distribution $ P_{\myVec{X}}$ via 
\begin{align}
&P_{X_i}(x_i) 
= \sum_{\{\myVec{x} / x_i\}}  P_{\myVec{X}}(\myVec{x}) 
\notag \\	&
= \underbrace{\left(\sum_{\{x_1, \ldots, x_{i-1}\}} \prod_{k=1}^{j} f_k(\mySet{V}_k) \right)}_{\triangleq \FwdMsg{f_{j}}{X_i}( x_i)} \underbrace{\left(\sum_{\{x_{i+1},\ldots, x_{\Blklen}\}} \prod_{k={j+1}}^{m} f_k(\mySet{V}_k) \right)}_{\triangleq \BwdMsg{f_{j+1}}{X_i}( x_i)}.
\label{eqn:MarginalFG}
\end{align}

The factorization of the joint distribution  implies that the marginal distribution, whose computation typically requires summation over $|\mySet{X}|^{\Blklen-1}$ variables, can now be evaluated as the product of two terms, $\FwdMsg{f_{j}}{X_i}( x_i)$ and   $\BwdMsg{f_{j+1}}{X_i}( x_i)$. These terms may be viewed as messages propagating forward and backward along the factor graph, e.g., $\FwdMsg{f_{j}}{X_i}( x_i)$ represents a forward message conveyed to edge $x_i$. In particular, these messages can be computed recursively. 
Writing $\mySet{V}_j = \{x_{i-\tau}, \ldots, x_{i}\}$ for some $\tau\ge 1$,  the \ac{sp} rule \cite{loeliger2004introduction} implies that
	\begin{align}
\FwdMsg{f_{j}}{X_i}( x_i) 
&= \sum_{\{x_{i-\tau},\ldots,x_{i-1}\}}   f_j(\mySet{V}_j) \sum_{\{x_1,\ldots,x_{i-\tau-1}\}} \prod_{k=1}^{j-1} f_i(\mySet{V}_i) \notag\\
	&=
 \sum_{\{x_{i-\tau},\ldots,x_{i-1}\}}   f_j(\mySet{V}_j) \FwdMsg{\tilde{f}_{i,j} }{X_{i-\tau}}( x_{i-\tau}),
\label{eqn:MsgPass}
	\end{align} 
 where the last equality follows from the fact that $x_{i-\tau}$ here is the variable with the largest index in $\mySet{V}_{j-1}$.
The computation of message terms in a recursive manner over a graphical model, as done in \eqref{eqn:MsgPass}, is referred to as {\em message passing}. In particular, the method of computing marginals in \eqref{eqn:MarginalFG} using message passing over factor graphs is referred to as the \ac{sp} algorithm \cite{kschischang2001factor}. 
\color{black}

\vspace{-0.2cm}
\section{Inference via Learned Factor Graphs}
\label{sec:Inference}
\vspace{-0.1cm} 	
In this section we present our proposed system for \ac{ml}-based inference applied to stationary time sequences. We begin by reviewing the application of the \ac{sp} method to stationary signals in Subsection \ref{subsec:SumProduct}. We next introduce the hybrid model-based/data-driven implementation of the \ac{sp} algorithm in Subsection \ref{subsec:BCJRNet} through the concept of learned factor graphs, which is followed by a discussion of this proposed method in Subsection \ref{subsec:Discussion}. 
Then, in Subsection \ref{subsec:ViterbiNet} we show how the learned factor graph can be used to carry out Viterbi detection, detailing how the architecture proposed in \cite{shlezinger2019viterbinet} for receiver design in digital communications can be obtained as a special case of a learned factor graph. 
Finally, we present training of learned factor graphs  in the presence of blockwise stationary distributions based on some future correctness indication in Subsection~\ref{subsec:OnlineTrain}.

\vspace{-0.2cm}
\subsection{\ac{sp} Inference for Stationary Markovian Time Sequences}
\label{subsec:SumProduct}
\vspace{-0.1cm}

\begin{figure}
	\centering
	{\includefig{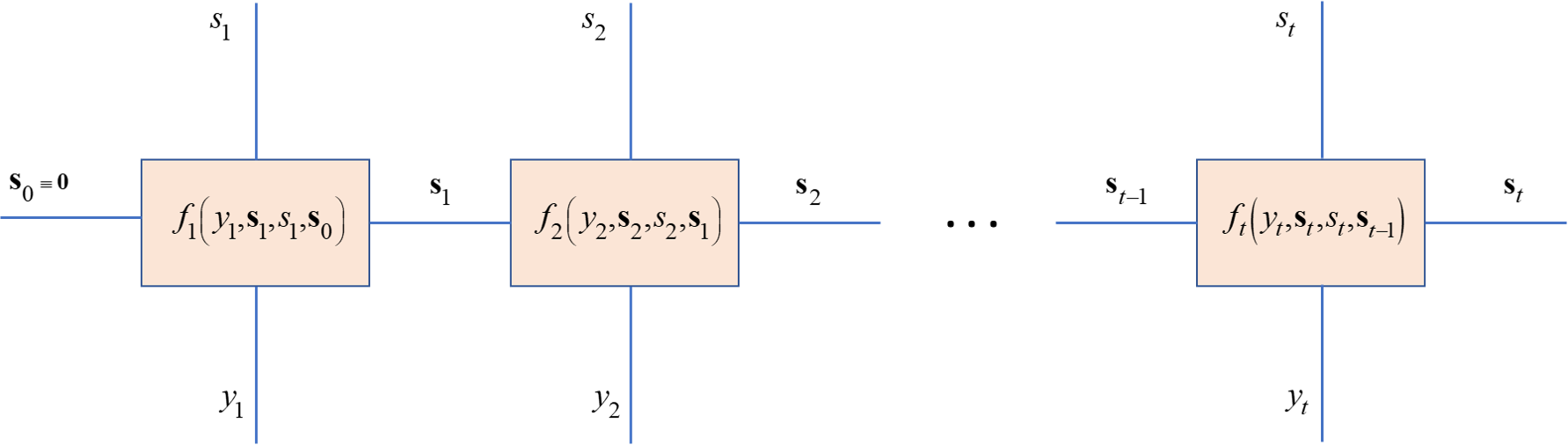}} 
	\caption{Factor graph of a Markovian time sequence.}
	\label{fig:FG_R2}	 
\end{figure} 

The \ac{sp} algorithm computes the \ac{map} rule in \eqref{eqn:MAP0} for the signal model detailed in Section \ref{sec:Model} by recursive message passing. To formulate this application of the \ac{sp} scheme, define the vector variable  $\myVec{s}_i \triangleq \myVec{s}_{i-\Mem+1}^i \in \mySet{S}^\Mem$ 
(similarly, the random vector  $\myVec{S}_i \triangleq \myVec{S}_{i-\Mem+1}^i \in \mySet{S}^\Mem$). We can now represent the factorizable joint distribution $\Pdf{\myVec{Y}^\Blklen,\myVec{S}^\Blklen}(\cdot)$ \eqref{eqn:MarkovModel} as the factor graph illustrated in Fig.~\ref{fig:FG_R2} (see \cite[Fig. 15]{loeliger2004introduction}), where the function nodes are
\begin{align*}
    f_i(y_i,\myVec{s}_i,s_i,\myVec{s}_{i-1}) \triangleq \Pdf{Y_i | \myVec{S}_{i-\Mem}^{i}}\!\left(y_i|\myVec{s}_{i-\Mem}^{i}\right)\! \Pdf{S_i |  \myVec{S}_{i-\Mem}^{i-1}}\!\left(s_i | \myVec{s}_{i-\Mem}^{i-1}\right).
\end{align*}
{Consequently, for the considered Markovian model, the partition sets $\{\mySet{V}_k\}$ defined in Subsection~\ref{subsec:FGModel} satisfy $\mySet{V}_k = \{ y_k. \myVec{s}_k, s_k, \myVec{s}_{k-1}\}$.} Due to the stationarity of the model, it holds that the mapping $f_i(\cdot)$ does not depend on the index $i$, and since $s_i$ is an element of the vector $\myVec{s}_i$, we can write the function nodes as 
\begin{align}
 f_i(y_i,\myVec{s}_i,s_i,\myVec{s}_{i-1}) &= 
\Pdf{Y_i | \myVec{S}_{i}, \myVec{S}_{i-1}}\left(y_i|\myVec{s}_{i}, \myVec{s}_{i-1}\right) \notag \\
&\times \Pdf{\myVec{S}_{i} |  \myVec{S}_{i-1}}\left(\myVec{s}_{i}| \myVec{s}_{i-1}\right) \notag \\
&\triangleq f\left({y}_i, \myVec{s}_i, \myVec{s}_{i-1} \right).
%
\label{eqn:FSC_funcNode}
\end{align}

\color{black}

\begin{figure}
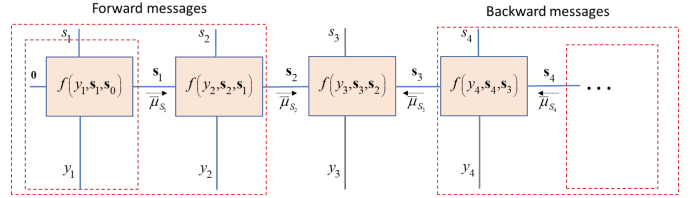

	\centering
	{\includefig{FG_State_R2.png}} 
	\caption{Message passing over the factor graph of a Markovian stationary time sequence.}
	\label{fig:SumProduct2}	 
\end{figure}

{In the special case where it holds that $\Pdf{Y_i | \myVec{S}_{i}, \myVec{S}_{i-1}}\left(y_i|\myVec{s}_{i}, \myVec{s}_{i-1}\right)  = \Pdf{Y_i | \myVec{S}_{i}, }\left(y_i|\myVec{s}_{i}\right)$ in \eqref{eqn:FSC_funcNode}, the statistical model and the corresponding factor graph coincides with that of a \acl{hmm}.}
Note that  when  $ \myVec{s}_i$ is a shifted version of   $\myVec{s}_{i-1}$,  \eqref{eqn:FSC_funcNode} coincides with  $\Pdf{Y_i | \myVec{S}_{i-\Mem}^{i}}\left(y_i|\myVec{s}_{i-\Mem}^{i}\right) \Pdf{S_i |  \myVec{S}_{i-1}}\left(s_i | \myVec{s}_{i-\Mem}^{i-1}\right)$,  otherwise it equals zero.   
The fact that the distribution is stationary implies that the function node mapping $f(\cdot)$ is invariant to the time index $i$. 
Using its factor graph representation, one can compute the joint distribution of $\myVec{S}^{\Blklen}$ and $\myVec{Y}^{\Blklen}$  by recursive message passing along its factor graph. In particular, 
\begin{align}
P_{\myVec{S}_k,\myVec{S}_{k+1}, \myVec{Y}^{\Blklen}}&({s}_k,{s}_{k+1}, \myVec{y}^{\Blklen}) = \notag \\
& \FwdMsg{f_k}{\myVec{S}_k}(\myVec{s}_k) f({y}_{k+1},   \myVec{s}_{k+1}, \myVec{s}_k) 
\BwdMsg{f_{k+2}}{\myVec{S}_{k+1}}(\myVec{s}_{k+1}),
\label{eqn:Recursion1}
\end{align}
where for $i = 1, \ldots, k$, the forward messages satisfy 
	\begin{equation}
\FwdMsg{f_i}{\myVec{S}_i}(\myVec{s}_i) = \sum_{\myVec{s}_{i-1}} f({y}_{i},  \myVec{s}_{i}, \myVec{s}_{i-1})\FwdMsg{f_{i-1}}{\myVec{S}_{i-1}}(\myVec{s}_{i-1}).
	\label{eqn:Recursion1Forwards}
	\end{equation}
Similarly, for $i = k+1 , \ldots, \Blklen-1$, the backward messages are 
	\begin{equation}
\BwdMsg{f_{i+1}}{ \myVec{S}_i}(\myVec{s}_i) = \sum_{\myVec{s}_{i+1}} f({y}_{i+1}, \myVec{s}_{i+1}, \myVec{s}_{i})\BwdMsg{f_{i+2} }{ \myVec{S}_{i+1}}(\myVec{s}_{i+1}).
	\label{eqn:Recursion1Backwards}
	\end{equation}
 This  message passing is  illustrated in Fig.~\ref{fig:SumProduct2}.

The ability to compute the joint distribution in \eqref{eqn:Recursion1} via message passing leads to computation of the \ac{map} detector in \eqref{eqn:MAP0} with complexity that grows linearly with $\Blklen$; without this message passing the computation    grows exponentially with the block size. This reduction in complexity is achieved by noting that the \ac{map} estimate satisfies
%
\begin{align}
\hat{s}_i\left( \myVec{y}^\Blklen\right)  
\!=\!\mathop{\arg \max}\limits_{s_i \in \mySet{S}} \sum_{  \myVec{s}_{i\! - \!1}\in \mySet{S}^{\Mem}}& \FwdMsg{f_{i\! - \!1}}{\myVec{S}_{i\! - \!1}}(\myVec{s}_{i\! - \!1}) f({y}_{i},  [s_{i\! - \!\Mem+1}, \ldots, s_{i} ],\myVec{s}_{i\! - \!1})  \notag \\
&\times \BwdMsg{f_{i+1}}{\myVec{S}_{i}}([s_{i\! - \!\Mem+1}, \ldots, s_{i} ]), 
\label{eqn:MAP2}
\end{align}
for each $i \in \Blkset$, where the summands can be computed recursively. When   $\Blklen$ is large, the messages may tend to zero. Hence, the messages are commonly scaled \cite{loeliger2004introduction}, e.g., $\BwdMsg{f_{i+1}}{\myVec{S}_i}(\myVec{s})$ is replaced with $\gamma_i \BwdMsg{f_{i+1}}{\myVec{S}_i}(\myVec{s})$ for some scale factor  that does not depend on $\myVec{s}$, and thus does not affect the \ac{map} rule. This instantiation of the \ac{sp} algorithm is summarized in Algorithm~\ref{alg:Algo0} below. 

\begin{algorithm}  
	\caption{ The \ac{sp} algorithm for stationary Markovian sequences}
	\label{alg:Algo0}
	\KwData{Fix an initial forward message $\FwdMsg{f_{0}}{\myVec{S}_{0}}(\myVec{s})= 1$ and a final backward message $\BwdMsg{f_{0}}{\myVec{S}_{\Blklen}}(\myVec{s})\equiv 1$. }
	\For{$i=\Blklen-1,\Blklen-2,\ldots,1$}{
		For each  $\myVec{s} \in \mySet{S}^\Mem$, compute backward message $ \BwdMsg{f_{i}}{\myVec{S}_{i}}(\myVec{s})$   via \eqref{eqn:Recursion1Backwards}
		\tcp*{backward messages}
	}
	\For{$i=1,2,\ldots,\Blklen$}{
		For each  $\myVec{s} \in \mySet{S}^\Mem$, compute forward message $ \FwdMsg{f_{i}}{\myVec{S}_{i}}(\myVec{s})$  via \eqref{eqn:Recursion1Forwards}
		\tcp*{forward messages}
	}
	\KwOut{$\hat{\myVec{s}}^\Blklen = [\hat{s}_1, \ldots, \hat{s}_\Blklen]^T$, each obtained using \eqref{eqn:MAP2}.}
\end{algorithm}
%
\vspace{-0.2cm}
\subsection{Learned Factor Graphs}
\label{subsec:BCJRNet}
\begin{figure}
	\centering
	{\includefig{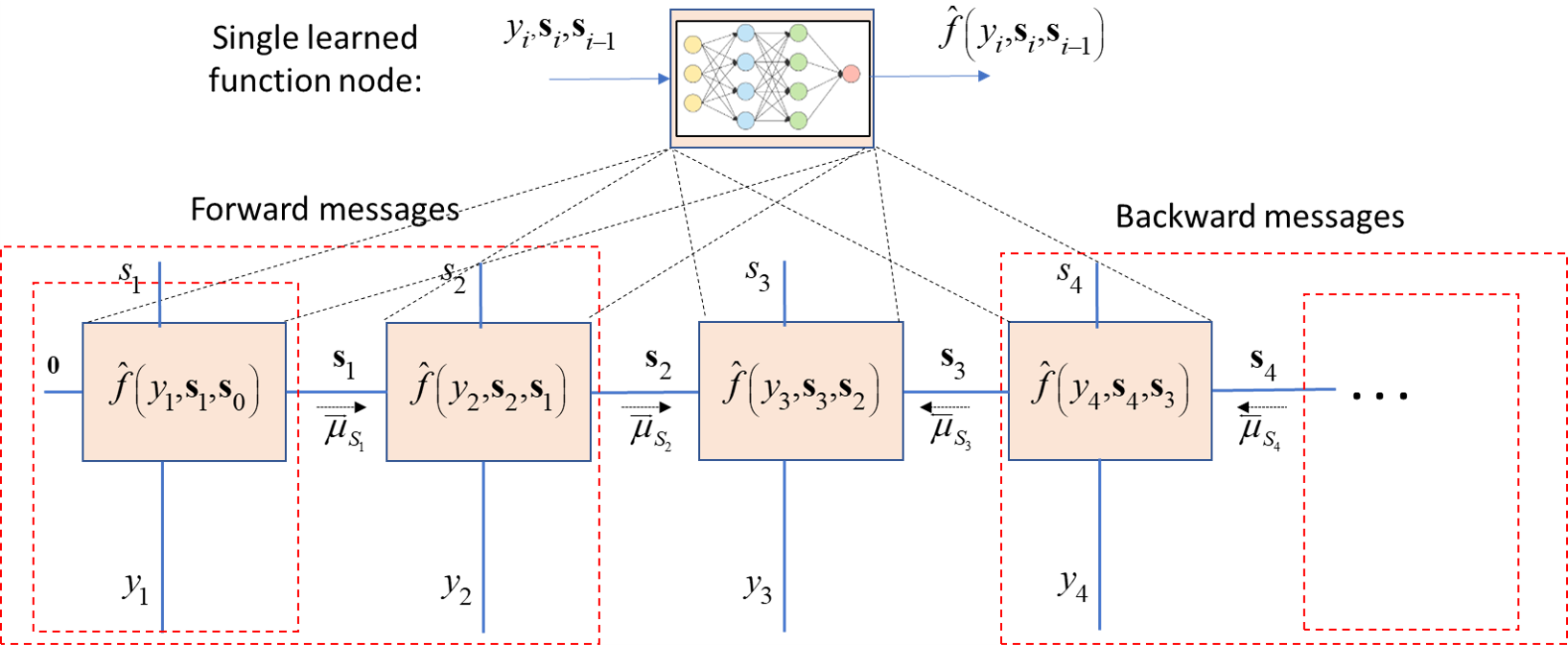}} 
	\caption{ \ac{sp} inference over a learned stationary factor graph.}
	\label{fig:LeanredSumProduct1}	 
\end{figure}

\vspace{-0.1cm}
Here, we propose a hybrid model-based/data-driven implementation of the \ac{sp} scheme in Algorithm~\ref{alg:Algo0}, which learns to implement \ac{map} detection of stationary Markovian time sequences from labeled data. 
Our framework builds upon the fact that in order to implement Algorithm~\ref{alg:Algo0}, one must be able to specify the factor graph representing the underlying distribution. In particular, the stationarity assumption implies that the complete factor graph is  encapsulated in the single function $f(\cdot)$ \eqref{eqn:FSC_funcNode} {\em regardless of the block size $\Blklen$}. The Markovian nature of the signals implies that the structure of the graph is known to be of the form detailed in the previous subsection and illustrated in Fig. \ref{fig:SumProduct2}, regardless of the actual values of its function nodes. Building upon this insight, we utilize \acp{dnn} to learn the mapping carried out at the function node separately from the inference task. 
By doing so,  one can train a system to learn an underlying factor graph, which can then be utilized for inference using conventional factor graph methods, such as the \ac{sp} algorithm.
The resulting learned stationary factor graph is then used to recover  $\{S_i\}$ by message passing, as illustrated in Fig. \ref{fig:LeanredSumProduct1}. 

In order to learn a stationary factor graph from samples, one must only learn its function node, which  boils down to learning  $P_{Y_i| \myVec{S}_{i-\Mem}^i }(\cdot)$ and $P_{S_i| \myVec{S}_{i-\Mem}^{i-1} }(\cdot)$ by \eqref{eqn:FSC_funcNode}.  Specifically, for stationary sequences, only a single function node must be learned, as the mapping $f(\cdot)$ does not depend on the time index $i$. 
This implies that one can utilize a single learned mapping, denoted $\hat{f}(\cdot)$, to carry out \ac{sp}-based inference over an arbitrary blocklength $\Blklen$, as illustrated in Fig. \ref{fig:LeanredSumProduct1}. 
When $\{S_i\}$ take values in a finite set, i.e., $\mySet{S}$ is finite, the transition probability $P_{S_i| \myVec{S}_{i-\Mem}^{i-1} }(\cdot)$  can be learned   via a histogram, as we do in our numerical study. For learning the  distribution $P_{Y_i| \myVec{S}_{i-\Mem}^i }(\cdot)$  
we consider two  architectures, based on classification and density estimation networks, respectively.

\subsubsection{Function Nodes as Classification Networks}   
Since $y_i$ is given and may take continuous values while the desired variables  take discrete values, a natural approach to evaluate $P_{Y_i| \myVec{S}_{i-\Mem}^i }(y_i | \myVec{s})$ for  each $\myVec{s}\in\mySet{S}^{\Mem+1}$   is to estimate $P_{\myVec{S}_{i-\Mem}^i |Y_i}(\myVec{s}|y_i)$, from which  $P_{Y_i| \myVec{S}_{i-\Mem}^i }(\cdot)$  is obtained using Bayes rule  as
	\begin{equation}
	\label{eqn:Bayes}
P_{Y_i| \myVec{S}_{i-\Mem}^i }(y_i | \myVec{s})  =
P_{\myVec{S}_{i-\Mem}^i |Y_i }\left(\myVec{s}|y_i \right) P_{Y_i  }\left(y_i   \right) {\big( P_{\myVec{S}_{i-\Mem}^i   }(\myVec{s} )\big)^{-1}}\!\!.
	\end{equation}
A parametric estimate of $P_{\myVec{S}_{i-\Mem}^i |Y_i }\left(\myVec{s}|y_i \right)$, denoted $\hat{P}_{\myVec{\theta}}(\myVec{s}|y_i)$,  is obtained  for each $\myVec{s}\in\mySet{S}^{\Mem+1}$ by training classification networks with softmax output layers to minimize the cross entropy loss. 

\begin{figure}
	\centering
	{\includegraphics[width=\figWidth]{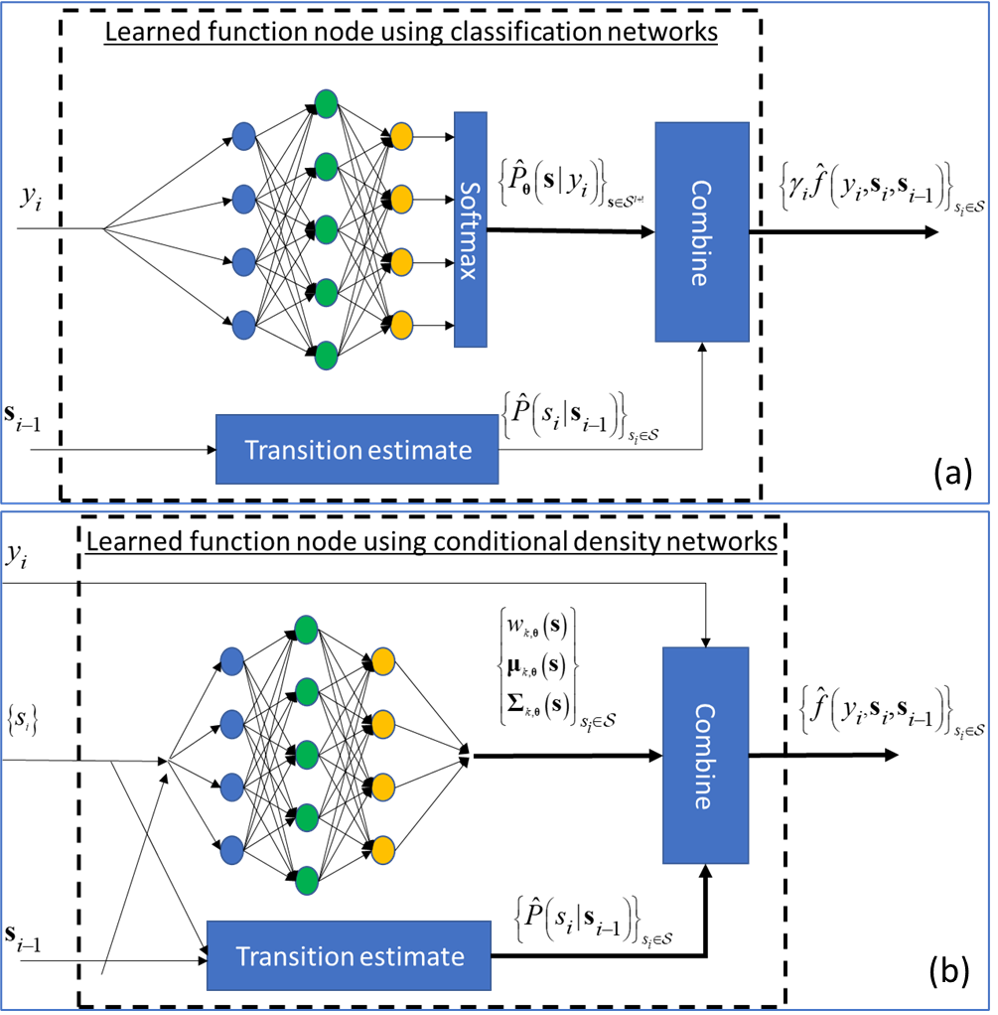}} 
	\caption{Learned function node architectures for evaluating $\hat{f}(\cdot)$ based on (a) classification \acp{dnn}, and (b) conditional density estimation networks.}
	\label{fig:LearnedFunctionNode}	 
\end{figure}

In general, the  marginal \ac{pdf} of $Y_i$ can be estimated from the training data using mixture density estimation via, e.g., \acl{em}  \cite[Ch. 2]{mclachlan2004finite}, or any other finite mixture model fitting method. 
However, while the joint distribution in \eqref{eqn:MarkovModel} depends on the marginal \ac{pdf} of $Y_i$, the \ac{map} rule is invariant of it. This follows since $ P_{Y_i  }\left(y_i   \right)$  does not depend on the variable $\myVec{s}$, and thus \eqref{eqn:MAP2} can be written as 
\begin{align}
\hat{s}_i\left( \myVec{y}^\Blklen\right)   
\!=\!\mathop{\arg \max}\limits_{s_i \in \mySet{S}} \sum_{  \myVec{s}_{i\! - \!1}\in \mySet{S}^{\Mem}}& \FwdMsg{f_{i\! - \!1}}{\myVec{S}_{i\! - \!1}}(\myVec{s}_{i\! - \!1}) \frac{f({y}_{i},  [s_{i\! - \!\Mem\!+\!1}, \ldots, s_{i} ],\myVec{s}_{i\! - \!1})}{ P_{Y_i }\left(y_i   \right)}  \notag\\
&\times \BwdMsg{f_{i+1}}{\myVec{S}_{i}}([s_{i\! - \!\Mem\!+\!1}, \ldots, s_{i} ]).
\label{eqn:MAP21}
\end{align}
Consequently, one can use a surrogate factor graph in which the function nodes are computed as $\frac{f({y}_{i},  \myVec{s}_{i} ],\myVec{s}_{i\! - \!1})}{ P_{Y_i }\left(y_i   \right)}$ instead of using \eqref{eqn:FSC_funcNode} without altering the predictions of the \ac{sp} algorithm. The surrogate function nodes, which differ from \eqref{eqn:FSC_funcNode} yet yield the same inference rule, are equivalently computed by setting  $ P_{Y_i }\left(y_i   \right) \equiv 1$ in \eqref{eqn:Bayes}.
\color{black}
%
The resulting structure  in which the  parametric  estimates are combined into a learned function node $\hat{f}(\cdot)$, scaled by some constant $\gamma_i = \frac{1}{P_{Y_i }\left(y_i   \right)}$, is illustrated in the upper part of Fig.~ \ref{fig:LearnedFunctionNode}.

\subsubsection{Function Nodes as Conditional Density Networks} 
An additional strategy is to directly estimate the conditional  $P_{Y_i| \myVec{S}_{i-\Mem}^i }\left(y_i | \myVec{s}  \right)$ from data. This can be achieved using conditional density estimation networks \cite{bishop1994mixture, rothfuss2019conditional}  that are specifically designed to learn such \acp{pdf}. Alternatively, normalizing flow networks \cite{rezende2015variational} can be used; these architectures which are typically used in the context of generative models, are capable of explicitly learning complex densities \cite{kobyzev2019normalizing}. For example, mixture density networks \cite{bishop1994mixture} model the conditional \ac{pdf} $P_{Y_i| \myVec{S}_{i-\Mem}^i }\left(y_i | \myVec{s}  \right)$ as a mixture of $K$ Gaussians, and train a \ac{dnn} to learn a parametric estimate of its mixing parameters, mean values, and covariances, denoted $w_{k,\myVec{\theta}}(\myVec{s} )$,  $\myVec{\mu}_{k,\myVec{\theta}}({\myVec{s}})$ and $\myMat{\Sigma}_{k,\myVec{\theta}}({\myVec{s} })$, respectively, by maximizing the likelihood of
\begin{equation}
\hat{P}_{\myVec{\theta}}\left(y_i | \myVec{s}  \right)= \sum_{k=1}^K w_{k,\myVec{\theta}}(\myVec{s} ) \mathcal{N}\big(y_i | \myVec{\mu}_{k,\myVec{\theta}}({\myVec{s} }), \myMat{\Sigma}_{k,\myVec{\theta}}({\myVec{s} })\big), 
\end{equation}
as illustrated in the lower part of Fig. \ref{fig:LearnedFunctionNode}.  


Both of the above architectures can be used for learning the conditional distribution $P_{Y_i| \myVec{S}_{i-\Mem}^i }(\cdot)$  utilized by the learned stationary factor graph. 
To provide guidelines for choosing between these architectures, we note that when  $\mySet{Y}$ is high-dimensional, directly learning the conditional density is difficult and likely to be inaccurate. In such cases, the classification-based architecture, which avoids the need to explicitly learn the density by accounting for the invariance of Algorithm \ref{alg:Algo0} to message scaling, may be preferable. When the state cardinality $|\mySet{S}|^{\Mem+1}$ is large, conditional density networks are expected to be more reliable. However, the \ac{sp} algorithm, which computes the messages for each possible state, becomes computationally infeasible when $|\mySet{S}|^{\Mem+1}$ grows, making the application of the \ac{sp} algorithm over learned factor graphs non-suitable for such setups. Consequently, in our numerical study  we use the classification network architecture for learning the function nodes.

\vspace{-0.2cm}
\subsection{Using Learned Factor Graphs for Inference}
\label{subsec:Discussion}
\vspace{-0.1cm}
The proposed approach of inference over learned factor graphs has several key advantages: 
First, as learning a single function node is expected to be a simpler task compared with learning the overall inference method for recovering $\myVec{S}^\Blklen$ from $\myVec{Y}^\Blklen$, this approach uses relatively compact \acp{dnn}, which can be learned  from a relatively small set of labeled data. Furthermore, the learned function node describes the factor graph for different values of $\Blklen$, implying that the same architecture can be used for inference from sequences with different lengths.  
When the learned function node is an accurate estimate of the true one, message passing over it effectively implements the \ac{map} detection rule \eqref{eqn:MAP0}, and thus approaches the minimal probability of error for each time instance.

Learned factor graphs rely on prior knowledge of the graph structure, which directly follows from the Markovian and stationarity assumptions.  As such, it incorporates this limited level of domain knowledge in the structure of the factor graph, but does not impose any assumptions on the function nodes. These properties allow the representation of complex and possibly analytically intractable joint distributions as learned factor graphs, as long as they obey the Markovian stationary structure.  {The mapping of the function nodes is learned in a model-invariant manner from labeled data comprised of realizations of the observed sequence along with the corresponding realizations of the hidden state. In practice, the latter can be obtained from measurements or based on human annotations, as in the numerical study detailed in Subsection~\ref{subsec:Sleep}.}
If additional domain knowledge is present, it can be incorporated via imposing some parametric model on the function nodes $\{ \hat{f}(\cdot)\}$, which follows from our understanding of the behavior of the setup at hand.  {Moreover, partial domain knowledge can be exploited to facilitate unsupervised training of the learned modules, thus relieving the dependence on known realizations of the state sequence, as recently proposed for \ac{dnn}-aided tracking based in \cite{revach2021kalmannet}. The operation of the proposed learned \ac{sp} inference is invariant of whether its \ac{dnn}-based function nodes were trained in a supervised or in an unsupervised manner. Nonetheless, we leave the study of unsupervised training of the learned function nodes for future investigation.}

The proposed strategy significantly simplifies inference for scenarios represented by cycle-free factor graphs, such as the considered Markovian setup, compared with previously proposed \acp{dnn} whose structure imitates the message passing operation trained end-to-end, such as factor graph neural networks \cite{zhang2019factor,satorras2020neural}. In particular, this approach enables the usage of compact \acp{dnn} which can be trained using small data sets and achieve improved performance over factor graph neural networks for cycle-free factor graphs, as numerically demonstrated in Section~\ref{sec:sims}.  Furthermore, the learned factor graph can be applied to stationary time sequences of different lengths, without having to change its architecture and train anew, as well as be utilized with different message passing mechanisms, as we show in Subsection~\ref{subsec:ViterbiNet}.  
For blockwise stationary sequences, one only needs to learn a different function node for each block, as discussed in Subsection  \ref{subsec:OnlineTrain}. Factor graph neural networks are expected to be advantageous for scenarios characterized by loopy factor graphs, i.e., setups not obeying the system model detailed in Subsection~\ref{subsec:FGModSequences}, where \ac{sp} inference does not coincide with the \ac{map} rule, and thus training end-to-end with sufficient data can lead to an improved inference rule.

{The proposed approach of inference over learned factor graphs bears some similarity to previous works on the optimization of information bounds for communication channels with memory \cite{sadeghi2009optimization, rusek2012bounds}. In particular, the channel model considered in \cite{sadeghi2009optimization} specializes to the stationary Markovian setup of \eqref{eqn:MarkovModel} under the additional constraint that the observations take values in a finite set. The main similarity to our work follows from the optimization over an auxiliary observations model, which \cite{sadeghi2009optimization} uses for formulating tight bounds on the information rates. Here, we use \acp{dnn} to capture the subtleties of this model from data such that the data-driven model can be integrated into factor graph based inference. Despite the similarity in the model and the approach, our work is somewhat different from these prior works on information rate bounds in the considered task as well as the usage of deep learning tools combined with principled model-based algorithms.}

The proposed approach for using learned factor graphs trains the function nodes separately from the inference algorithm which utilizes the factor graph. Consequently, the  factor graph learned can  be processed using various message passing algorithms other than the \ac{sp} scheme, e.g., the max-product method \cite{loeliger2004introduction} and the Viterbi algorithm \cite{viterbi1967error}. In particular, in the following subsection we  show that the function nodes of learned factor graphs assuming  equiprobable $\{S_i\}$ produce the same learned quantities as that used by ViterbiNet, proposed in \cite{shlezinger2019viterbinet} for symbol detection in finite-memory communication channels.
{Furthermore, the rationale of inference over learned factor graphs can be combined with alternative factor graph structures and message passing mechanisms, such as those used for joint channel estimation and iterative detection in  \cite{NovakTSP2013,RieglerTIT2013,HansenTSP2018a,KirkelundGlobeCom2010}.}
Finally, we expect the same design to be  applicable when training in an end-to-end manner, i.e., by backpropating through the  message passing algorithm, as was done in \cite{knobelreiter2020belief} for image segmentation problems. We leave these extensions for future research. 

\vspace{-0.2cm}
\subsection{Application as the Viterbi Algorithm}
\label{subsec:ViterbiNet}
\vspace{-0.1cm}	
\ac{sp} inference over learned factor graphs detailed in Subsection \ref{subsec:BCJRNet} is based on the ability to learn a parametric estimate of the function node mapping for factor graphs of stationary time sequences. {However, once the factor graph is learned from data, it can also be used by factor graph inference algorithms other than the \ac{sp} method.  One such alternative designed for Markovian observations is the Viterbi scheme \cite{viterbi1967error}, originally proposed for decoding convolutional channel codes in digital communications. While the  Viterbi algorithm can be applied with both forward and backward recursions over the underlying factor graph,  we focus here on its implementation with a single forward recursive computation, following the description of the Viterbi scheme in \cite[Ch. 3.4]{tse2005fundamentals} and \cite[Ch. 8.3]{goldsmith2005wireless}.}

\subsubsection{{The Viterbi Algorithm}}
The Viterbi detector aims at recovering the maximum likelihood sequence estimator in \eqref{eqn:ML0} via
\begin{align}
\hat{\myVec{s}}^{ \Blklen}\left( \myVec{y}^{ \Blklen}\right)  
&= \mathop{\arg \min}_{\myVec{s}^{ \Blklen} \in \mySet{S}^\Blklen } -\log \Pdf{\myVec{Y}^{ \Blklen} | \myVec{S}^{ \Blklen}}\left( \myVec{y}^{ \Blklen} | \myVec{s}^{ \Blklen}\right)  \notag \\
&= \mathop{\arg \min}_{\myVec{s}^{ \Blklen} \in \mySet{S}^\Blklen }\sum\limits_{i=1}^{\Blklen } - \log \Pdf{Y_i  | \myVec{S}_i}\left( y_i  | \myVec{s}_i\right),
\label{eqn:ML3}
\end{align}
where in \eqref{eqn:ML3} we use the abbreviated term  {$\myVec{S}_i$ defined in Subsection~\ref{subsec:SumProduct}, i.e., $\myVec{S}_i = \myVec{S}_{i-\Mem+1}^{i}$}. 
The optimization problem \eqref{eqn:ML3} can be solved recursively using dynamic programming, by  iteratively updating a {\em path cost} $c_i(\myVec{s})$ for each state $\myVec{s}\in \mySet{S}^\Mem$. The resulting scheme, known as the Viterbi algorithm, is given below as Algorithm~\ref{alg:Algo1}. 
\begin{algorithm}  
	\caption{ The Viterbi Algorithm \cite{viterbi1967error}}
	\label{alg:Algo1}
	\KwData{Fix an initial path $\myVec{p}_{0}\left(\myVec{s}\right) = \varnothing $ and path cost ${c}_{0}\!\left(\myVec{s}\right)\! =\!  0$, for each  $\myVec{s} \in \mySet{S}^\Mem$. }
	\For{$i=1,2,\ldots,\Blklen$}{
		For each  $\myVec{s} \in \mySet{S}^\Mem$, compute previous state with shortest path, denoted $\myVec{u}_\myVec{s}$, via 
		\begin{equation}
		\label{eqn:ViterbiUpdate}
		\myVec{u}_\myVec{s} \!=\! \!\mathop{\arg\min}\limits_{\myVec{u} \in \mySet{S}^\Mem:  \Pdf{\myVec{S}_i|\myVec{S}_{i-1}}\!(\myVec{s}|\myVec{u})>0 }\! \left({c}_{i-1}\!\left(\myVec{u}\right)  - \log \Pdf{Y_i  | \myVec{S}_i}\!\left( y_i  | \myVec{s}\right) \right).
		\end{equation}\\
		Update cost and path via 			
		\begin{equation}
		\label{eqn:ViterbiUpdate2}
		{c}_i\left( \myVec{s}\right) = {c}_{i-1}\left(\myVec{u}_\myVec{s} \right)  - \log \Pdf{Y_i  | \myVec{S}_i}\left( y_i  | \myVec{s}\right), 
		\end{equation}
		and $\myVec{p}_{i}\left(\myVec{s}\right)  = \big[\myVec{p}_{i-1}\left(\myVec{u}_\myVec{s}\right) , \myVec{u}_\myVec{s} \big]$\;
	}
	\KwOut{$\hat{\myVec{s}}^\Blklen = \myVec{p}_{\Blklen}\left(\myVec{s}^*\right)$ where $\myVec{s}^* = \arg\min_\myVec{s} c_\Blklen(\myVec{s})$.}
\end{algorithm}

%
\textcolor{NewColor}{Algorithm~\ref{alg:Algo1} outputs its estimate of  the complete unknown sequence, i.e.,  $\hat{\myVec{s}}^\Blklen$, at time instance $\Blklen$. However, its output can also be approximated in real-time, since all paths at time $i+\Mem$, $\{\myVec{p}_{i+\Mem}(\myVec{s})\}_{\myVec{s}\in\mySet{S}^\Mem}$ typically include the same states corresponding to time instances not larger then $i$ \cite{forney1973viterbi,lou1995implementing}. Consequently,  the output of Algorithm \ref{alg:Algo1} can be approached by estimating $\hat{\myVec{s}}_i$  when processing $y_{i+\Mem}$, and  thus produce its estimates with a constant delay of $\Mem$ time instances. It is emphasized though that while this implementation typically comes at a negligible performance loss compared to Algorithm~\ref{alg:Algo1}, it is no longer guaranteed to recover the maximum likelihood sequence estimator.}

\subsubsection{Viterbi Detection over Learned Factor Graphs}
To see that the learned factor graph detailed in Subsection~\ref{subsec:BCJRNet} can also be applied for Viterbi detection, we focus on the case where \textcolor{NewColor}{the conditional distribution of $\myVec{S}_i$ given $\myVec{S}_{i-1}$ is uniform}, as is commonly the case in digital communications systems for which Algorithm~\ref{alg:Algo1} was originally derived. 
\color{black}
In such cases the shortest path equation \eqref{eqn:ViterbiUpdate} can be written as 
\begin{align}
\myVec{u}_\myVec{s} &= \mathop{\arg\min}\limits_{\myVec{u}\in  \mySet{S}^\Mem} \left({c}_{i-1}\left(\myVec{u}\right)  \!- \!\log \Pdf{Y_i  | \myVec{S}_i}\left( y_i  | \myVec{s}\right)\! -\! \log \Pdf{\myVec{S}_i|\myVec{S}_{i-1}}(\myVec{s}|\myVec{u})\right) \notag \\
&= \mathop{\arg\min}\limits_{\myVec{u} \in \mySet{S}^\Mem} \left({c}_{i-1}\left(\myVec{u}\right)  - \log f_i\left( y_i , \myVec{s}, \myVec{u}\right) \right),
\label{eqn:ViterbiUpdateA}
\end{align}	
where $f_i(\cdot)$ is defined in \eqref{eqn:FSC_funcNode}.

Similarly, the cost update equation \eqref{eqn:ViterbiUpdate2} can be replaced with 
\begin{equation}
\label{eqn:ViterbiUpdate2a}
{c}_i\left( \myVec{s}\right) = {c}_{i-1}\left(\myVec{u}_\myVec{s}\right)  - \log f_i\left( y_i , \myVec{s}, \myVec{u}_\myVec{s}\right), 
\end{equation}	
without affecting the resulting inference rule. 
%
Furthermore, scaling the function nodes  $f_i(\cdot)$ by some $\gamma_i$ that is independent of the state does not affect the Viterbi algorithm due to the $\arg \min$ statements in  Algorithm~\ref{alg:Algo1}. This implies that the same learned factor graph proposed in Subsection~\ref{subsec:BCJRNet} for \ac{sp} inference, which trains parametric estimates of $f_i(\cdot)$, can be utilized to also carry out Viterbi detection in a data-driven manner.   In fact, carrying out the Viterbi algorithm (rather than the \ac{sp} method) over the learned factor graphs coincides with ViterbiNet, proposed in \cite{shlezinger2019viterbinet} for symbol detection in finite-memory communications. The application of the model-based \ac{sp} algorithm in such scenarios, i.e., symbol detection in finite-memory channels, specializes to the BCJR algorithm \cite{bahl1974optimal}. Thus \ac{sp} inference over learned factor graphs of the joint input-output distribution of finite memory communications implements BCJR detection from data.

\subsubsection{Discussion}
Learned factor graphs can thus be applied, once trained, to carry out multiple inference algorithms, including the \ac{sp} scheme (as proposed in Subsection~\ref{subsec:BCJRNet}) as well as the Viterbi algorithm (via ViterbiNet). 
{Furthermore, while Algorithm~\ref{alg:Algo1} considers the combination of learned factor graphs with Viterbi detection to produce hard decisions, where the output is the vector $\hat{\myVec{s}}^{\Blklen}\in \mySet{S}^{\Blklen}$, similar computations can be used to output soft decisions by utilizing the soft-output Viterbi algorithm over the learned factor factor graph.}
Once the factor graph encapsulating the underlying distribution is learned, one can decide which  inference algorithm to apply to a learned factor graph. The preference of one method over the other is invariant of the learned factor graph, and follows from the differences between model-based message passing schemes, e.g., the differences between the \ac{sp} algorithm and the Viterbi algorithm.
The main advantages of Algorithm~\ref{alg:Algo1} over  Algorithm~\ref{alg:Algo0}, and thus of using a learned factor graph as part of ViterbiNet over \ac{sp} inference, are its reduced complexity and \textcolor{NewColor}{the fact that it can be approached using a} real-time operation. In particular, while the complexity of both algorithms grows linearly with the block size $\Blklen$, the Viterbi scheme as detailed in Algorithm~\ref{alg:Algo1} computes only a forward recursion and can thus provide its estimations in real time within a given delay from each incoming observation, while the \ac{sp} scheme implements both forward and backward recursions, and can thus infer only once the complete block is observed. One can also implement the \ac{sp} method using only the forward messages, and thus share the real-time operation and reduced complexity of Algorithm~\ref{alg:Algo1}, at the cost of reduced accuracy and deviation from the \ac{map} rule.

The main advantage of  Algorithm~\ref{alg:Algo0} over  Algorithm~\ref{alg:Algo1}, i.e., of using the learned factor graphs for \ac{sp} inference with forward and backward recursions rather than as part of the ViterbiNet system, stems from the fact that it implements the \ac{map} rule \eqref{eqn:MAP0}, which minimizes the symbol error probability. The maximum likelihood sequence detector \eqref{eqn:ML0} computed by the Viterbi algorithm, requires the states to be equiprobable in order to be able to  approach the performance of the \ac{map} rule. In digital communications, where  the Viterbi algorithm originates, the states correspond to transmitted symbols, which are commonly equiprobable, and thus the Viterbi detector is far more popular and widely used compared with the \ac{sp} scheme. However, in many other problems involving inference from time sequences, such as sleep pattern detection considered in Subsection \ref{subsec:Sleep}, the states do not obey a uniform distribution, making the combination of learned factor graphs with \ac{sp} inference the more attractive and natural data-driven method for such tasks.

\vspace{-0.2cm}
\subsection{Application for Blockwise-Stationary Statistical Variations}
\label{subsec:OnlineTrain}
\vspace{-0.1cm}	
In the previous subsections we discussed how one can learn to infer from time sequences by using  \acp{dnn} for estimating the function nodes in factor graphs with known structures, rather than using these networks for the complete end-to-end inference task. As a result, inference over learned factor graphs can be carried out using relatively compact networks which are trainable with a small number of training samples, as we also numerically demonstrate in our experimental study detailed in Section~\ref{sec:sims}. This property of  learned factor graphs facilitates their operation in the presence of blockwise stationary distributions. 

As detailed in Subsection \ref{subsec:FGModSequences}, under a blockwise stationary distribution, the conditional \ac{pdf} $\Pdf{Y_i  | \myVec{S}_i}\left(\cdot\right) $, which is the mapping produced by a neural network in learned factor graphs, changes every  $\NonStatBlk$ time instances. Such scenarios correspond to, e.g., communication over block-fading channels. 
In general, one can tackle these time variations by joint learning \cite{oshea2017introduction,xia2020note}, i.e., training the function nodes using labeled data corresponding to a broad set of expected distributions. However, this approach requires a large data set, and the resulting mapping may be inaccurate for each of the observed   $\Pdf{Y_i  | \myVec{S}_i}\left(\cdot\right)$. Alternatively, one can train in advance a different network for each statistical block via ensemble models \cite{raviv2020data}.  Such a strategy is feasible when one has prior access to training data corresponding to each statistical relationship as well as knowledge regarding the order in which these statistical models are observed during inference. An additional approach involves the usage of meta-learning tools as in \cite{park2020meta, raviv2021meta} for tuning the architecture and training hyperparameters of the data-driven function node architecture such that it can be rapidly re-trained to accurately represent a given family of expected distributions.

Finally, when some future indication on the inference correctness is available, the ability to train data-driven function nodes individually using a small number of labeled samples can be exploited to track temporal variations in the underlying statistics, without requiring prior knowledge of these variations. This is achieved by using the future indication to re-train the network in a self-supervised manner. For example, in a digital communications setup, the desired $\myVec{S}^\Blklen$ represents the transmitted symbols, while the observed $\myVec{Y}^\Blklen$ is the output of the channel used for recovering these symbols. Such communications are typically protected using error correction coding, implying that even when some of the symbols are inaccurately estimated, the conveyed message is still recoverable, as long as the number of errors does not exceed the code distance \cite[Ch. 8]{goldsmith2005wireless}. In such cases, the recovered message can be re-encoded, generating the postulated transmitted symbols, which in turn can be used along with corresponding observations for re-training the learned parameters, as proposed in \cite{shlezinger2019viterbinet, teng2020syndrome, shlezinger2019deepSIC}. When this network can be effectively adapted using a small training set of the order of a communication codeword, the resulting data-driven digital communication receiver is capable of tracking the variations of the underlying statistical model, thereby avoiding errors caused by these variations. We numerically demonstrate the gains of this approach for adapting learned factor graphs to track blockwise variations in the context of digital communications in our experimental study described in Subsection \ref{subsec:symdet}.


\vspace{-0.2cm}
\section{Experimental Results}
\label{sec:sims}
\vspace{-0.1cm}		
We next numerically evaluate inference over learned factor graphs in an experimental study.  
First,  we consider the problem of sleep pattern detection, using the PhysioNet Sleep-EDF Expanded database \cite{kemp2013sleep}. Then, we focus on symbol detection in communication over finite-memory channels, using simulated data of common channel models. {We conclude the section with an evaluation of the computational complexity associated with inference via learned factor graphs.} 
Throughout this section, we use the classification network architecture (upper part of Fig. \ref{fig:LearnedFunctionNode}), trained with the Adam optimizer \cite{kingma2014adam} to minimize the cross-entropy loss  for learning the function nodes.\footnote{\label{ftn:Github} The source code used in this section is available online in the following link: \url{https://github.com/nirshlezinger1/LearnedFactorGraphs}} 
The state transition probability is estimated from the training data provided in each experiment via a histogram, i.e., for $n_t$ labeled training samples $\{s_k,y_k\}_{k=1}^{n_t}$, it is computed as
	\begin{equation}
	    \hat{P}(s_i|\myVec{s}_{i-1}) \!=\! \frac{\sum_{k=\Mem+1}^{n_t} \mathds{1}\left([\myVec{s}_{i-1}^T, s_i ]\! =\! [s_{k-\Mem}, \ldots, s_k]\right)}{\sum_{k=\Mem+1}^{n_t} \mathds{1}\left(\myVec{s}_{i-1}^T = [s_{k-\Mem}, \ldots, s_{k-1}]\right)},
	    \label{eqn:TransMat}
	\end{equation}
		where $\mathds{1}(\cdot)$ is the indicator function.
\color{black}
We utilize relatively compact networks which require only several minutes to train on a standard CPU. The specific architectures used for each setup are detailed in the description of the corresponding scenarios below.

	\vspace{-0.2cm}
\subsection{Sleep Pattern Detection}
	\label{subsec:Sleep}
	\vspace{-0.1cm}
Here, we consider the problem of sleep pattern detection from EEG signals. We use the PhysioNet Sleep-EDF Expanded database \cite{kemp2013sleep}, which consists of 197 whole-night PolySomnoGraphic sleep recordings, containing EEG, EOG, chin EMG, and event markers. Similar to many prior works in this area \cite{supratak2017deepsleepnet, phan2018automatic, humayun2019end}, we use $20$ patients from one of the two studies in this dataset that investigates the age effect in healthy subjects, known as the Sleep Cassette (SC) dataset. We  focus on using a single EEG channel recording (the EEG channel Fpz-Cz) to classify five stages of sleep: awake (AWA), REM, and non-REM sleep stages (N1-N3), i.e., $	|\mySet{S}| = 5$. In particular, every 30 seconds of recording (i.e., 3000 EEG samples at 100 Hz), which is called an epoch, is labeled by human experts. For some of our experiments, we apply the feature extraction method proposed in \cite{jiang2019robust} to extract $150$ features  in each epoch, used as the observation $Y_i$.

The task here is to identify the sleep states from the observed EEG signal. The common strategy  is to train highly-parameterized \acp{dnn} to predict $S_i$ from $Y_i$, based on convolutional layers \cite{supratak2017deepsleepnet, humayun2019end} or bidirectional \acp{rnn} \cite{phan2018automatic}.  
Our goal here is to show that by using the compact networks associated with learned factor graphs, one can achieve comparable performance to previously proposed deep detectors. These compact networks can be trained using smaller data sets and are simpler to implement compared with previously proposed \acp{dnn}. We also show that the learned factor graph framework can be used to improve upon existing architectures, using their predictors for learning the function nodes instead of for directly recovering the desired $\{S_i\}$. 
 
%

	\begin{figure} 
		\centering
		{\includegraphics[height=1.1in]{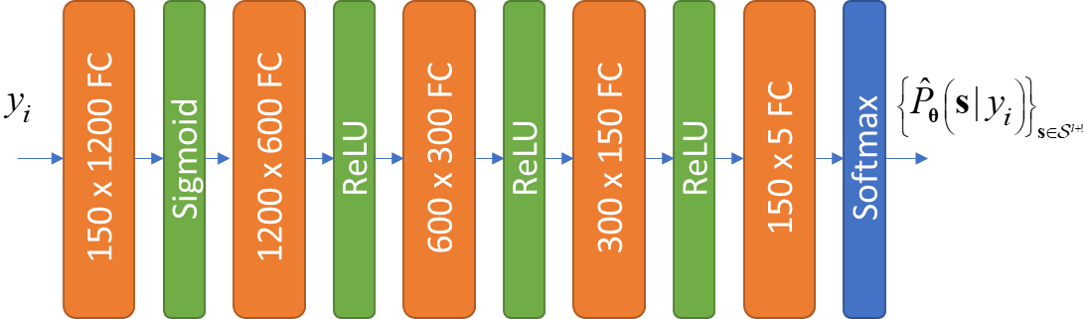}}
		\caption{The 5 FC network used for learning the function nodes.
		}
		\label{fig:5FC_Architecture}  
	\end{figure}

We learn the factor graph assuming that the sleep states follow a first-order Markov chain model, i.e., $\Mem = 1$, and the joint distribution of states and measurements is stationary.  In particular, we utilize a five layer \ac{fc} network (referred to as 5 FC)  consisting of: $150\times 1200$ \ac{fc} layer with sigmoid activation;   $1200 \times 600$,   $600 \times 300$, and  $300 \times 150$ \ac{fc}  layers with ReLU activations; and $150 \times 5$ \ac{fc} layer with softmax output layer. An illustration of this network is depicted in Fig. \ref{fig:5FC_Architecture}. Here we use the 150 features extracted from EEG signals at each epoch as the network input. We also use a $34$ layer residual network (referred to as 34 ResNet) proposed in \cite{humayun2019end}, where the network input is the raw EEG signals during each epoch. For both of these networks, we compare the  accuracy of the trained network directly applied for sleep pattern detection with that achieved when  used as learned function nodes for \ac{sp} inference.    

\begin{figure}				
	\centering 
	\begin{subfigure}[b]{\figWidth}
		\centering
		\includegraphics[width=\figWidth]{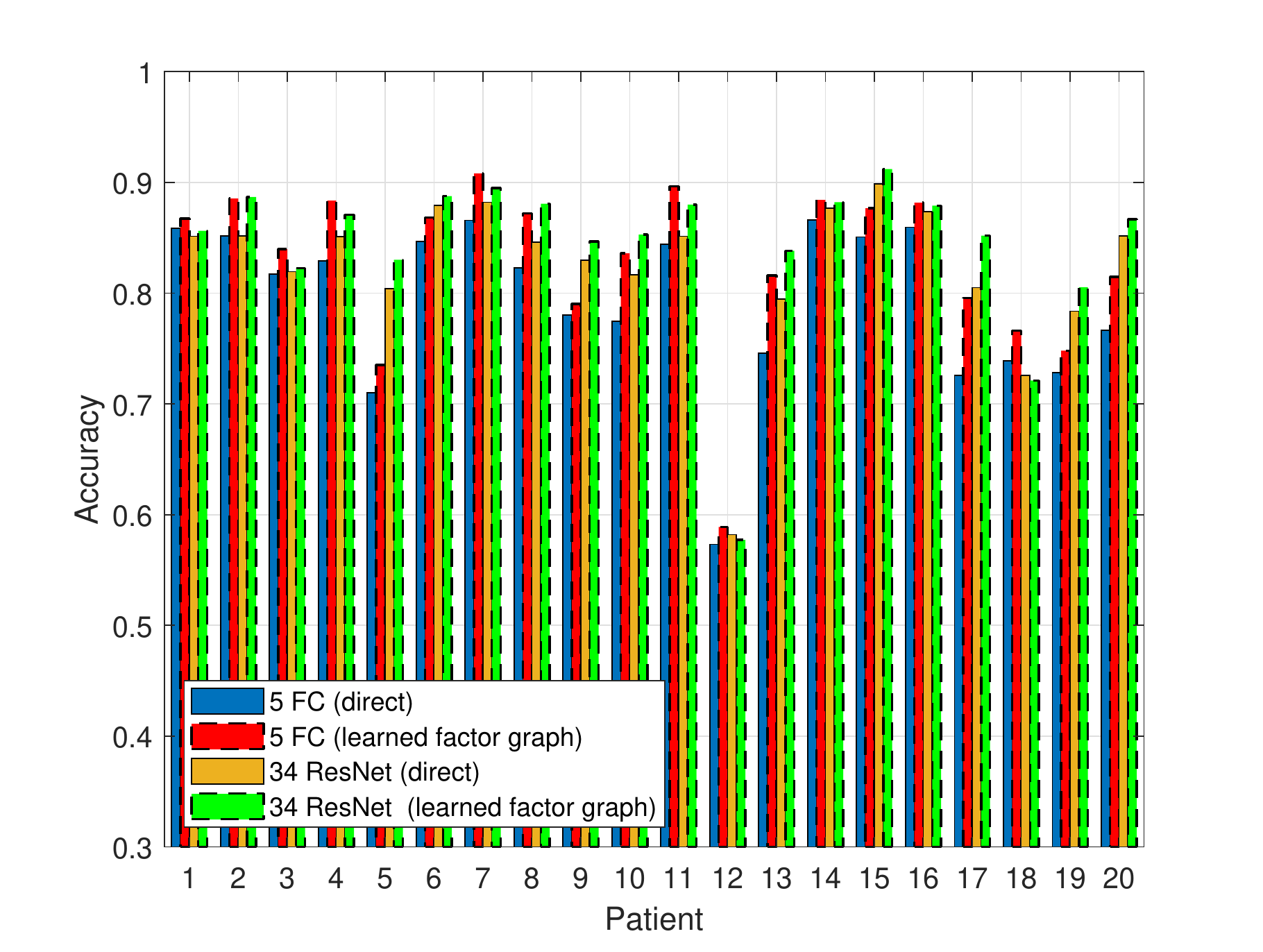}
		\caption{LOO cross validation accuracy.}\label{fig:LOO_5FC}
	\end{subfigure}

	\begin{subfigure}[b]{\figWidth}
		\centering
		\includegraphics[width=\figWidth]{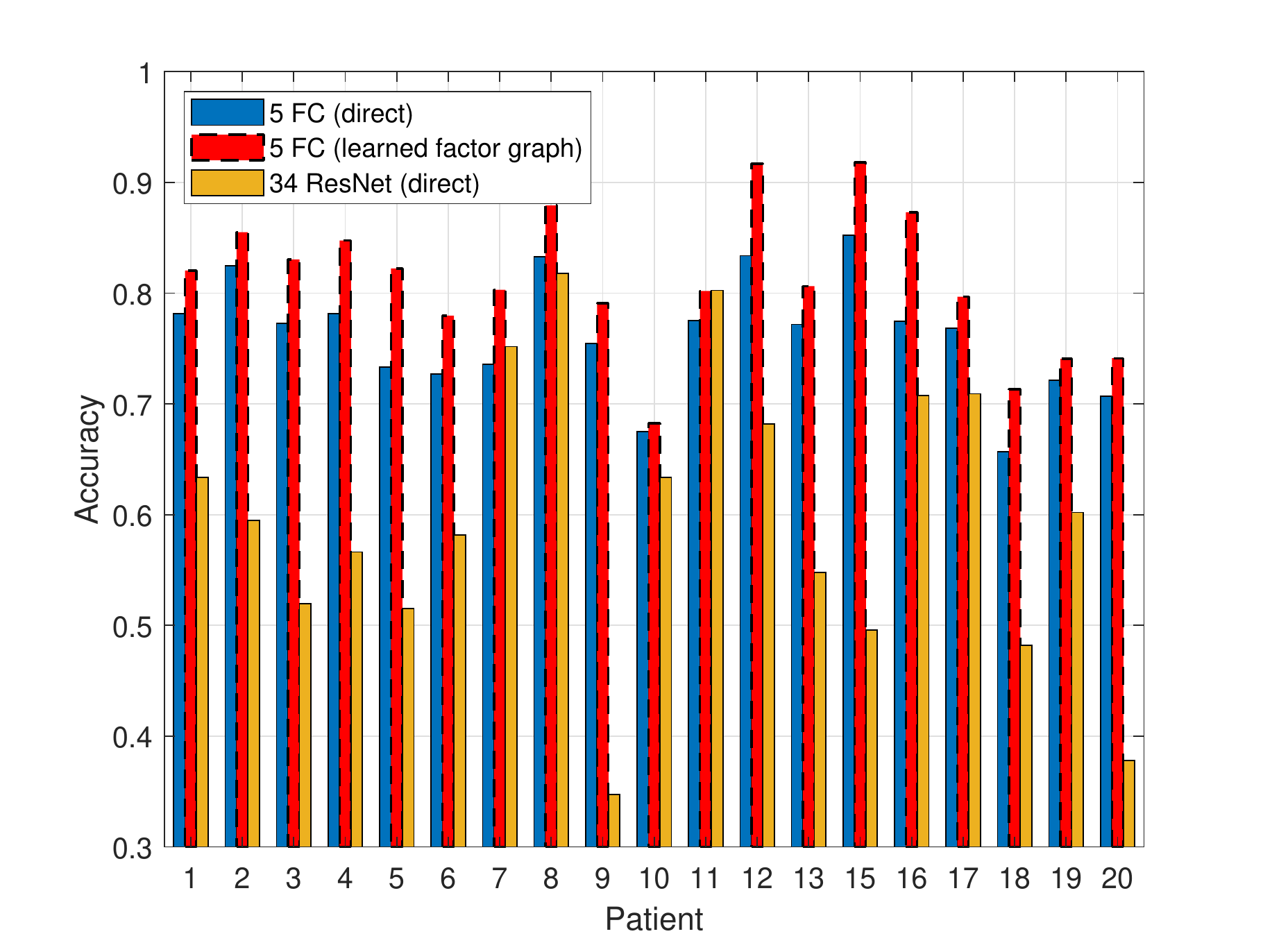}
		\caption{Train over $1000$ samples accuracy.}\label{fig:TrainOn1000samples_5FC}
	\end{subfigure} 
	\caption{Sleep pattern detection numerical results.}
	\label{fig:Sleep}
\end{figure}

\begin{figure*}				
	\centering
			\begin{subfigure}[b]{.3\linewidth}
				\includegraphics[width=\linewidth, height=1.5in]{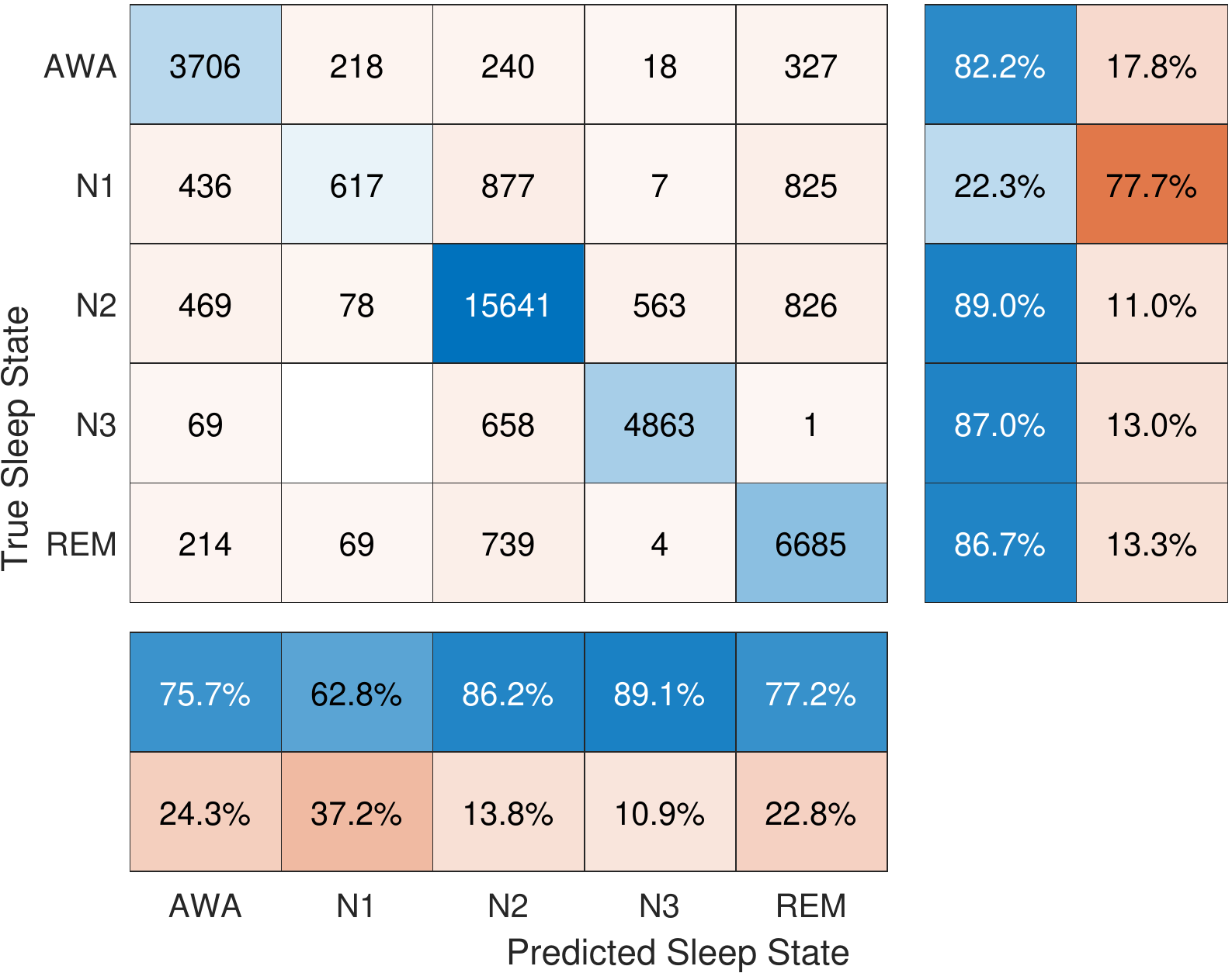}
				\caption{5 FC (LOO).}\label{fig:LOO_5FC_conf}
			\end{subfigure} 
			$\quad$
			\begin{subfigure}[b]{.3\linewidth}
				\includegraphics[width=\linewidth, height=1.5in]{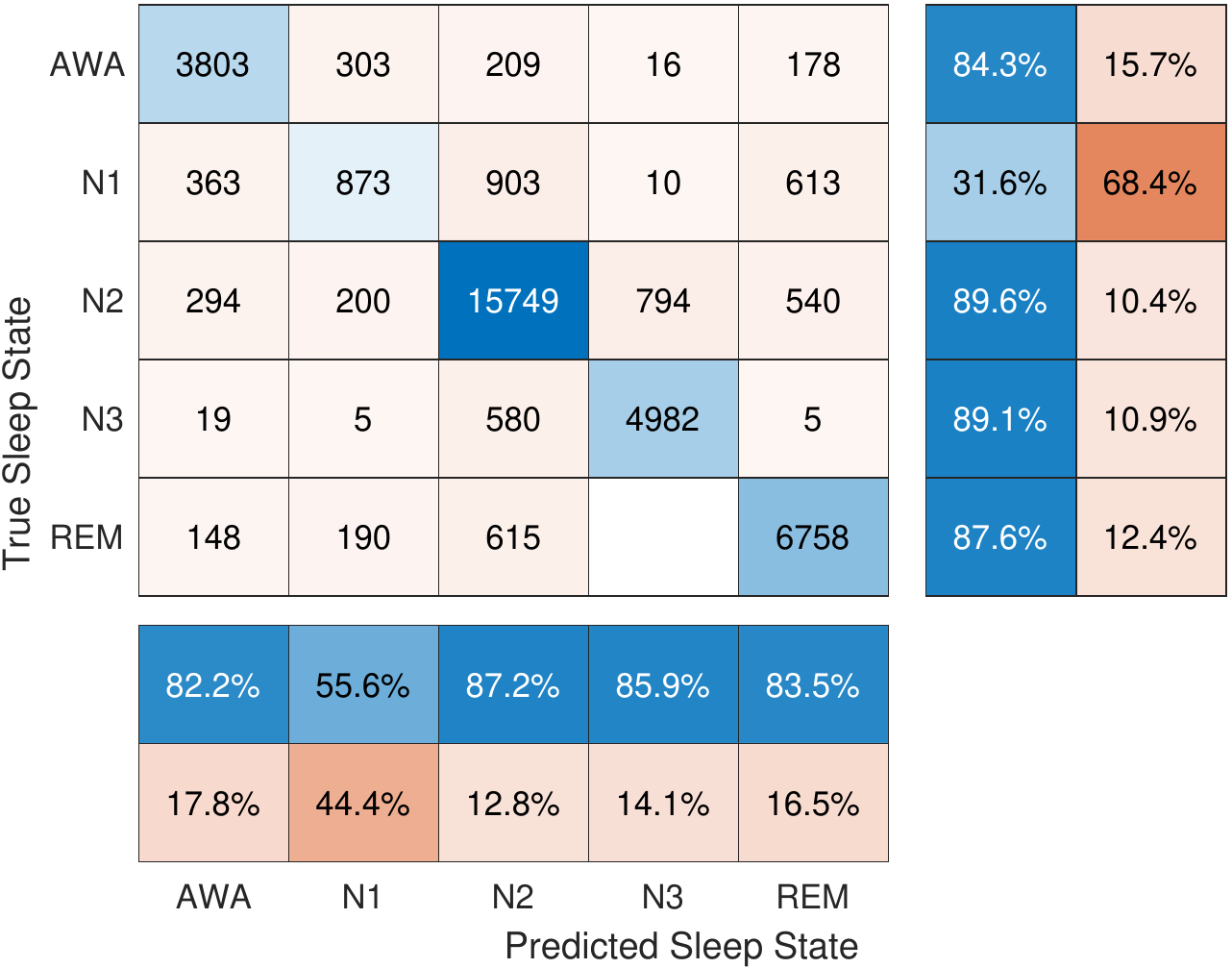}
				\caption{34 ResNet    (LOO).}\label{fig:LOO_ResNet_conf}
			\end{subfigure} 
			$\quad$
			\begin{subfigure}[b]{.32\linewidth}
				\includegraphics[width=\linewidth, height=1.5in] {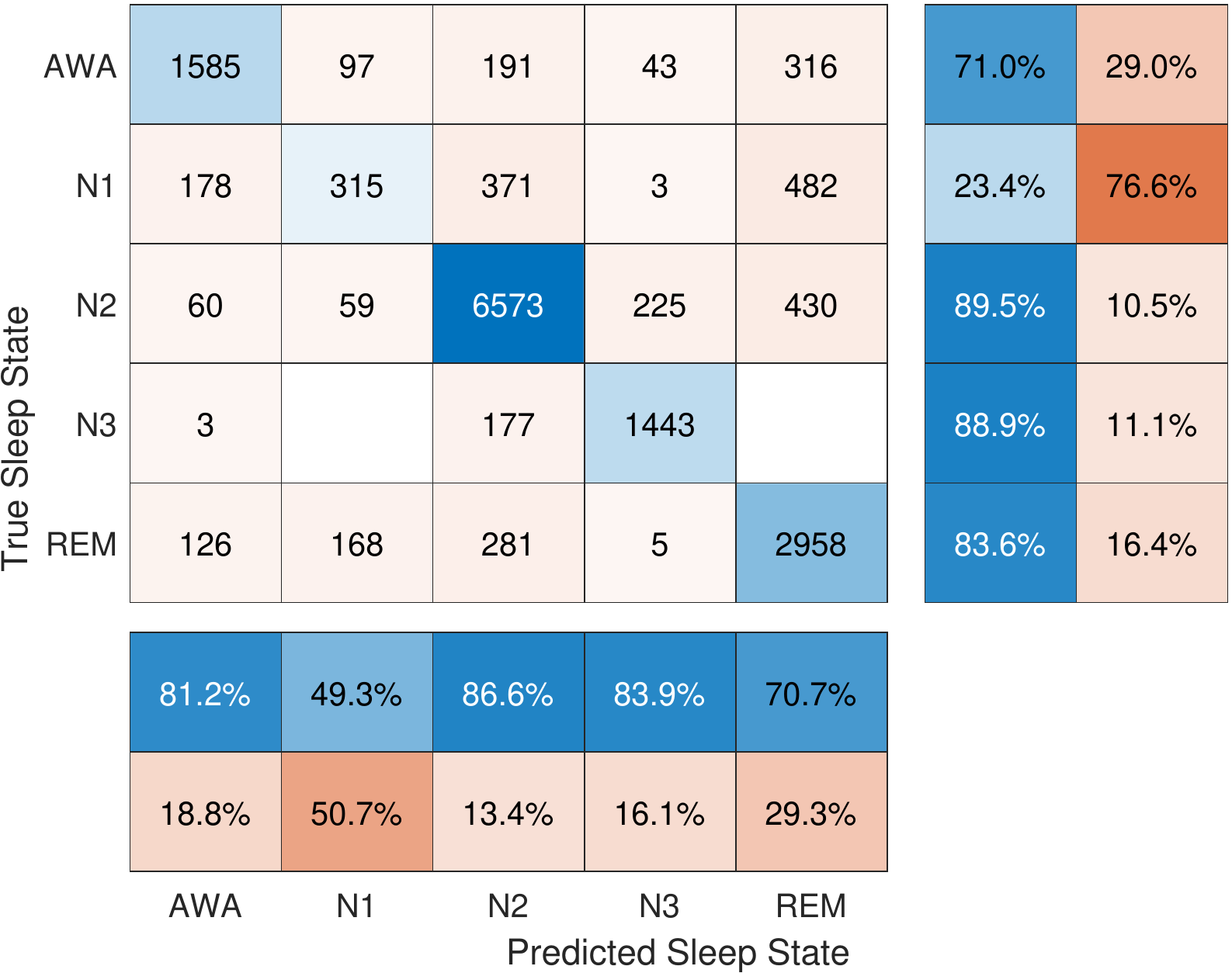}
				\caption{5 FC   (Train on 1000 samples).} 
				\label{fig:TrainOn1000samples_conf}
			\end{subfigure}	 
	\caption{Sleep pattern detection confusion matrices of \ac{sp} inference over learned factor graphs.}
	\label{fig:SleepConf} 
	\vspace{-0.2cm}
\end{figure*}

First, we apply \ac{loo} cross-validation, where the data from one patient is left out for evaluation, while the data from the $19$ other patients is used for training the models.  
\begin{table} 
	\centering
	\vspace{-0.2cm}
	\begin{tabular}[r]{|p{5.5cm}|p{1.7cm}|}
		
		\hline
		Method & Accuracy \\ \hline\hline 
		CNN-BLSTM \cite{supratak2017deepsleepnet} & 82.0\% \\
		BLSTM-SVM \cite{phan2018automatic} & 82.5\% \\
		5 FC (direct)  & 79.2\% \\ 
		5 FC (learned factor graph) & 82.8\% \\
		34 ResNet (direct)  & 82.3\% \\
		34 ResNet (learned factor graph) & {\bf 84.2\%} \\
		
		\hline
	\end{tabular} 
	\caption{\ac{loo} overall accuracy.}\label{tab: Accuracy} 
\end{table}
The average accuracy of the algorithms for each patient is shown in Fig. \ref{fig:LOO_5FC}, while the confusion matrices achieved using learned factor graphs when using the 5 \ac{fc} network and with 34 ResNet are shown in Figs. \ref{fig:LOO_5FC_conf}-\ref{fig:LOO_ResNet_conf}, respectively. Here, the 5 \ac{fc} network is trained with an initial learning rate of $0.001$ over $50$ epochs with a mini-batch size of $60$ samples. Directly applying 5 FC achieves an average accuracy of $79.2\%$, while using it to learn the factor graph in \ac{sp} inference achieves an  improvement of  $3.6\%$, resulting in an  accuracy of $82.8\%$. Similarly, the 34 ResNet achieves an accuracy of $82.3\%$ while applying it to learn the factor graph achieves an average improvement of about $2\%$, resulting in an accuracy of $84.2\%$. As summarized by Table~\ref{tab: Accuracy}, when using either the 5 \ac{fc} network or the 34 ResNet  classifier  to learn the factor graph, our algorithm outperforms the state-of-the-art deep learning algorithms applied to this dataset \cite{supratak2017deepsleepnet, phan2018automatic}.


We also observe in Fig. \ref{fig:LOO_5FC} that for some patients, such as patient $12$, the samples appear to obey a  considerably different statistical model from that of the remaining patients. 
\ifFullVersion
This degrades the classification accuracy of the compact network, which in turn leads to an inaccurate estimate of the function nodes,  resulting in accuracy below $75\%$. 	
This  motivates us to evaluate learned factor graphs when trained and tested using samples from the same patient, exploiting the compact networks of then learned factor graphs, which facilitates training from small data sets. We thus compute  the accuracy when, for each patient, the networks are trained using its first $1000$ samples and tested using the remaining samples, except for patient $14$ for which less than $1000$ samples are available. 

The average accuracy for each patient is depicted in Fig. \ref{fig:TrainOn1000samples_5FC}, and the confusion matrix achieved by learned factor graphs using the 5 \ac{fc} network for learning the function nodes is shown in Fig. \ref{fig:TrainOn1000samples_conf}. Here, directly applying the 5 FC network as a classifier achieves an accuracy of $76\%$, as the \ac{dnn} is trained using only $1000$ samples, using it to form a learned factor graph improves the accuracy to $81\%$.  34 ResNet achieves only $60\%$ accuracy due to its inability to properly train its highly parameterized network using small datasets.
%
Building upon the ability of the relatively compact network to adapt with few samples,  learned factor graphs  achieve improved accuracy when applied to patients whose measurements obey a unique statistical model. For example,  the \ac{sp} method over the learned factor graph achieves an accuracy of $92\%$ when applied to patient $12$, improving by over $8\%$ compared with using the network as predictor, and by $25\%$,  compared with its performance when trained over the remaining $19$ patients in Fig. \ref{fig:LOO_5FC}. 
These results demonstrate the potential of combining neural networks for learning the function nodes  rather than to carry out the complete classification tasks, as well as the advantages of this approach in allowing the usage of compact networks, which can be trained with  small training sets, for accurate inference from time sequences with non-synthetic data.

\vspace{-0.2cm}
 \subsection{Symbol Detection in Digital Communications}
	\label{subsec:symdet}
	\vspace{-0.1cm}	
Next, we apply learned factor graphs for detection in a finite-memory communication setup. We use  simulated data based on common  channel models, which allows us to compare the performance of the \ac{sp} method over learned factor graphs with that of the model-based \ac{sp} algorithm, as well as to that of existing deep detectors. 	Here, a transmitter sends a sequence of symbols $S_i \in \mySet{S}$, $i \in \Blkset$, and a receiver uses the channel output $Y_i \in \mathbb{R}$ to recover the symbols.  
Each $Y_i$ is affected only by the last $\Mem$ transmitted symbols, where $\Mem$ is the  memory length.

We consider two channels with memory length $\Mem=4$: A Gaussian channel  and a Poisson channel. 
Let $\myVec{h} (\gamma)\in \mathbb{R}^\Mem$ be a vector whose entries obey an exponentially decaying profile  $h_\tau(\gamma) \triangleq e^{-\gamma(\tau-1)}$ for $\gamma > 0$ and $\tau \in \{1,\ldots, \Mem\}$. For the Gaussian channel, the symbols {take values in the set $\mySet{S} = \{-1, 1\}$, representing a binary phase shift keying constellation,}  and the channel output is generated via
	\begin{equation}
	\label{eqn:AWGNCh1}
Y_i | \myVec{S}^\Blklen \sim \mathcal{N} \left(\sqrt{\rho} \cdot\sum_{\tau=1}^{\Mem}h_\tau(\gamma) S_{i-\tau + 1}, 1\right),
	\end{equation}
where $\rho > 0$ represents the \ac{snr}.  
	For the Poisson channel, the channel input represents on-off keying, i.e., $\mySet{S} = \{0,1\}$, and the channel output $Y_i$ obeys
	\begin{equation}
	\label{eqn:PoissonCh1}
Y_i | \myVec{S}^\Blklen\sim \mathcal{P}\left( \sqrt{\rho} \cdot\sum_{\tau=1}^{\Mem}  h_\tau(\gamma) S_{i-\tau + 1} + 1\right),
	\end{equation}
	where $\mathcal{P}(\cdot)$ is the Poisson distribution. {Namely, for each realization $\myVec{S}^\Blklen = \myVec{s}^\Blklen$, the channel output for every time instance is generated independently, where the output at corresponding to the $i$th time instance is generated from a Poisson distribution with parameter which equals $\sqrt{\rho} \cdot\sum_{\tau=1}^{\Mem}  h_\tau(\gamma) s_{i-\tau + 1} + 1$ \cite{farsad2018neural}.}

We implement the classification network  with three \ac{fc} layers:  
$1 \times 100$, $100 \times 50$, and  $50 \times  16$ layers, using  sigmoid and ReLU activation functions, respectively.  
The network is trained using $5000$  samples, which is the order of a  typical preamble sequence in wireless networks \cite{dahlman20103g}, with learning rate $0.01$. The training is carried out over  $100$ epochs with mini-batch size of $27$. {Using these training samples to compute the transition probability via \eqref{eqn:TransMat} yields an estimate with normalized mean-squared error of merely $1.01\cdot 10^{-4}$ with respect to the true transition probability which equals $\frac{1}{|\mySet{S}|}=\frac{1}{2}$ here.} The \ac{dnn} architecture is depicted in Fig. \ref{fig:3FC_Architecture}.

	\begin{figure} 
	\centering
	{\includegraphics[height=1.1in]{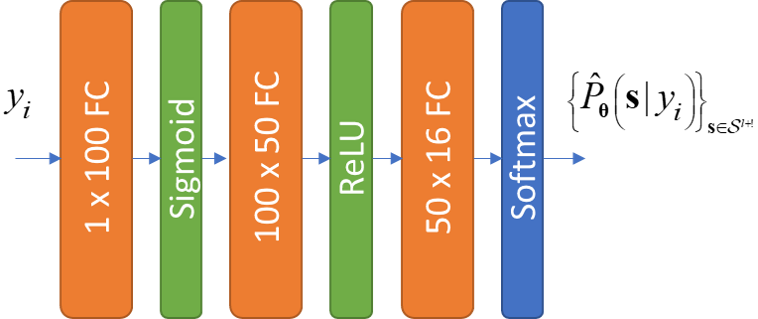}}
	\caption{The 3 FC network used for learning the function nodes.
	}
	\label{fig:3FC_Architecture} 
\end{figure}

For each channel, we  compute the \ac{ser} achieved using learned factor graphs for  different values of the \ac{snr}  $\rho$, and the \ac{dnn}-aided function node is trained anew for each value of $\rho$.
For every \ac{snr}, the \ac{ser} values are averaged over $20$ different channel vectors  $\myVec{h} (\gamma)$, obtained by letting $\gamma$ vary in the range $[0.1, 2]$. 
For comparison, we  evaluate the \ac{ser} of the model-based \ac{sp} algorithm, as well as that of the data-driven  \ac{sbrnn} deep  detector proposed in \cite{farsad2018neural}
{and the factor graph neural network of \cite{zhang2019factor}, for which we use the same architecture as that utilized in the numerical study in \cite[Sec. 4.1]{zhang2019factor}. }
{We consider two cases: The first is {\em perfect \ac{csi}}, in which the \ac{sp} method knows the exact $\myVec{h}(\gamma)$, while the data-driven systems are trained using data consisting of samples generated with the same  $\myVec{h}(\gamma)$ used for  the test data. The second setup considered, referred to as {\em \ac{csi} uncertainty},  considers the scenario in which the \ac{sp} algorithm is implemented using an estimate of $\myVec{h}(\gamma)$ corrupted by Gaussian noise with variance which equals $10\%$ and $8\%$ of the magnitude of the channel tap for the Gaussian  and Poisson channels, respectively. Using the considered setup for \ac{csi} uncertainty represents the operation of the model-based \ac{sp} algorithm in scenarios where the underlying model is not accurately known, in a manner that is invariant of how this uncertainty is obtained. For the \ac{dnn}-aided systems, the training data is generated with  the noisy  $\myVec{h}(\gamma)$, allowing us to study  resiliency  to inaccurate training.}
In all cases, the information symbols are uniformly randomized in an i.i.d. fashion from $\mySet{S}$, and 
the test samples are generated from their corresponding channel 
with the true   vector $\myVec{h}(\gamma)$.

\begin{figure}
	\centering
	\vspace{-0.4cm}
	\begin{subfigure}[b]{\figWidth}
		\centering
		\includegraphics[width=\figWidth]{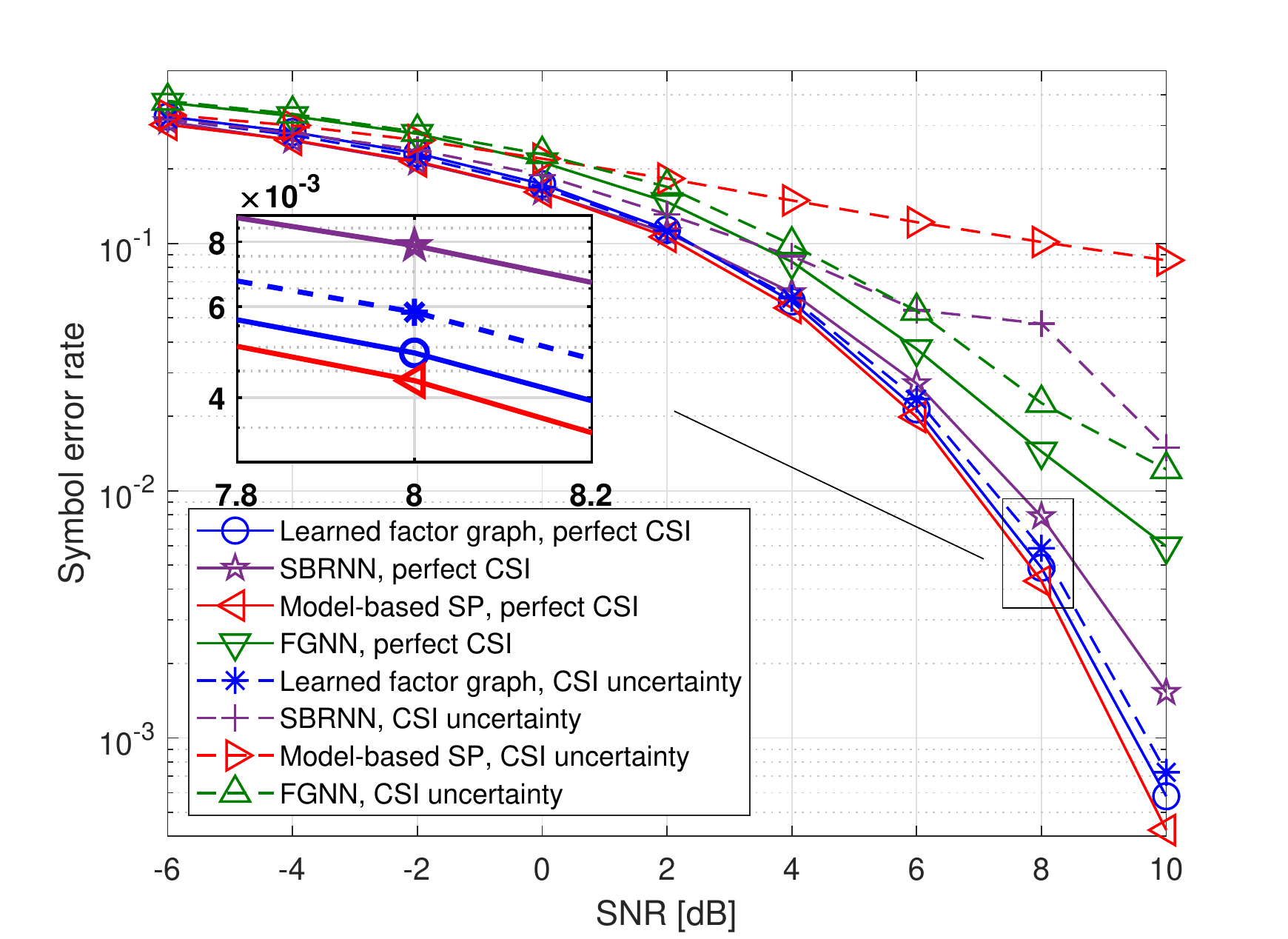}
		\caption{SER performance of different detectors in the Gaussian channel.}\label{fig:AWGN}
	\end{subfigure}
	
	\begin{subfigure}[b]{\figWidth}
		\centering
		\includegraphics[width=\figWidth]{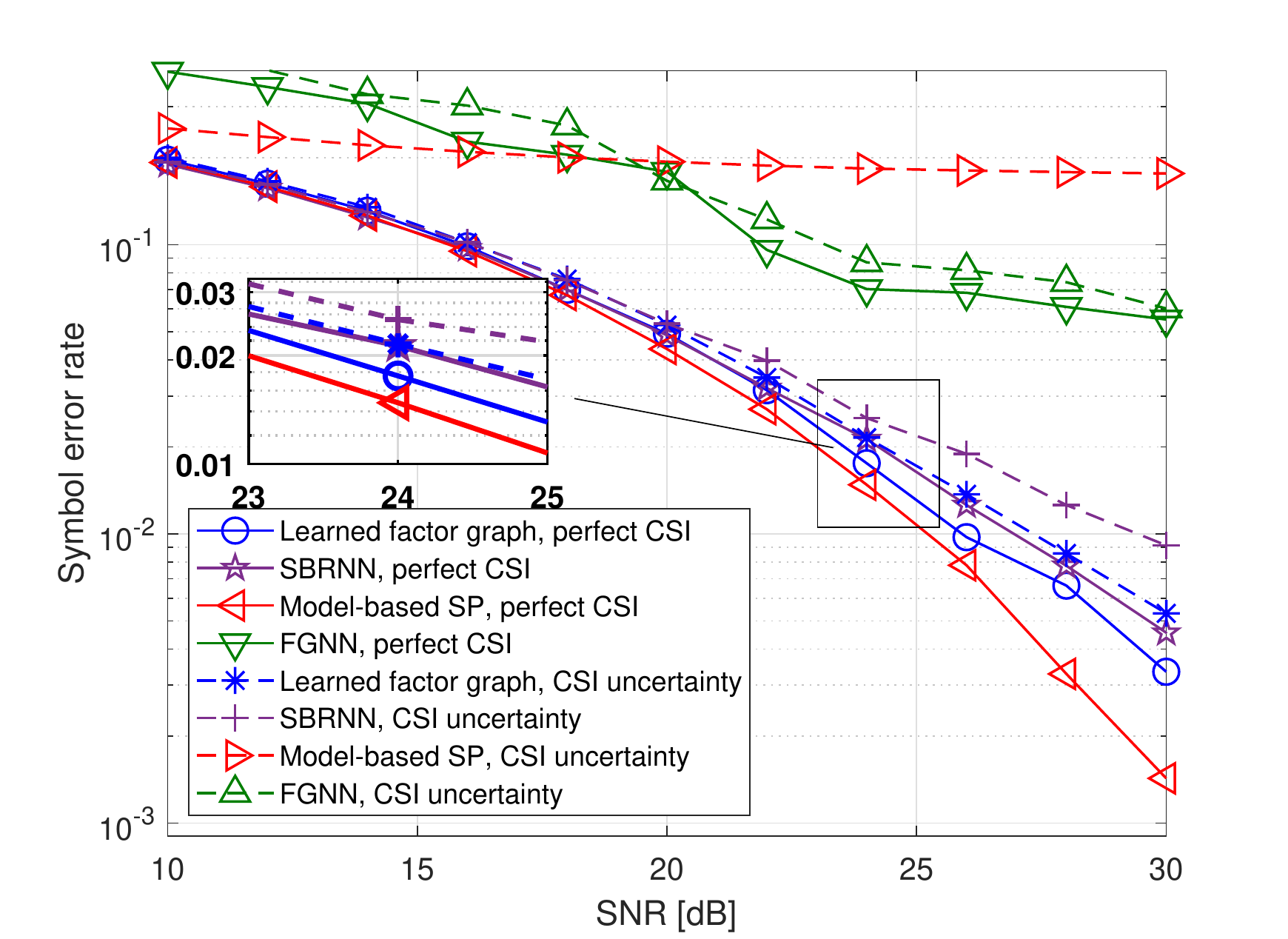}
		\caption{SER performance of different detectors in the Poisson channel.}\label{fig:Poisson}
	\end{subfigure} 
	\caption{Symbol detection accuracy of the \ac{sp} algorithm over learned factor graphs compared the data-driven SBRNN detector \cite{farsad2018neural} and the purely model-based \ac{sp} method.}	 
	\label{fig:Comm}
\end{figure}	

The  \ac{ser} values, averaged over $50000$ Monte Carlo simulations,  are depicted in Figs. \ref{fig:AWGN}-\ref{fig:Poisson} for the Gaussian and Poisson channels, respectively. We observe in Figs. \ref{fig:AWGN}-\ref{fig:Poisson} that the \ac{ser} achieved using learned factor graphs approaches that of the \ac{sp} algorithm from which it originates, while the latter requires accurate prior knowledge of the underlying distribution.
In the presence of \ac{csi} uncertainty,  carrying out \ac{sp} inference over a learned factor graph significantly outperforms applying it over the inaccurate model-based factor graphs. When the function nodes are trained with a variety of different channels, learned factor graphs achieve relatively good \ac{ser}  when inferring under each of the channels for which it is trained, while the performance of the conventional \ac{sp} method is significantly degraded due to imperfect \ac{csi}. 
We also observe that the \ac{sbrnn} receiver,  shown in \cite{farsad2018neural} to approach the performance of the \ac{map} rule when sufficient training is provided,
is outperformed by \ac{sp} inference over learned factor graphs here due to the small training set.  
{A similar observation is noted for the factor graph neural network of \cite{zhang2019factor}, whose \ac{ser} performance is within a notable gap from  by the proposed approach of learned factor graphs for the considered scenario.}
These results demonstrate the ability of learned factor graphs  to enable accurate implementation of the \ac{sp} method while requiring small training sets and improving robustness to  uncertainty.

Next, we consider blockwise stationary channels, showing how the approach discussed in Subsection \ref{subsec:OnlineTrain} can exploit coded communications for channel tracking. Here, transmission consists of multiple codewords of length $\NonStatBlk = 2040$, each representing $1784$ bits encoded using a  \ac{rs} [255, 223] channel code, and protected with a \ac{crc} for error detection. We simulate the Gaussian and Poisson channels in \eqref{eqn:AWGNCh1}-\eqref{eqn:PoissonCh1}, respectively. To simulate block-wise temporal variations, we let the entries of  $\myVec{h}(\gamma)$ vary between codewords. As in \cite{shlezinger2019viterbinet}, for the $j$th codeword, we use  $h_{\tau}(\gamma) \triangleq e^{-\gamma(\tau-1)}\cdot\Big(0.8 + 0.2\cos\big( \frac{2\pi \cdot j}{p_\tau}\big)  \Big)$ for each $\tau \in \{1,\ldots, \Mem\}$, {with a fixed exponential decay parameter $\gamma = 0.2$}, and with $ \myVec{p} = [51, 39, 33, 21]^T$, representing block-wise periodic variations in the channel  coefficients. 
 
Before the first block is transmitted, a factor graph representing a stationary distribution is learned using $5000$ training samples taken using the initial channel coefficients. In order to track channel variations via online training, we use successful decoding to re-train the function nodes in a decision-directed manner \cite{shlezinger2019viterbinet,teng2020syndrome}. In particular, each recovered block $\hat{\myVec{S}}^{\NonStatBlk}$ is decoded to its corresponding $1784$ bits using an \ac{rs} decoder, and validated using a \ac{crc} check. If the \ac{crc} check passes, the bits are re-encoded using an \ac{rs} encoder into a postulated symbol block  $\tilde{\myVec{S}}^{\NonStatBlk}$, which is used along with its corresponding observed $\myVec{Y}^{\NonStatBlk}$ to retrain the learned function node using $50$ epochs with an initial learning rate of $0.002$. 

In addition to evaluating learned factor graphs with online training, we also compute the coded \ac{ber}  when the function node is trained only once using the $5000$ training samples representing the initial channel, referred to as {\em initial training}, as well as when trained once using $5000$ training samples corresponding to the channels observed at blocks $j \in 3\cdot\{1,\ldots,10\}$, referred to as {\em joint training}. The coded \ac{ber} of \ac{sp} inference over learned factor graphs is compared with that of the SP detector with full instantaneous \ac{csi} as well as to that with knowledge of only the initial channel conditions. The coded \ac{ber} results, averaged over $200$ consecutive blocks, are depicted in Figs. \ref{fig:TVChannelAWGN}-\ref{fig:TVChannelPoisson} for the Gaussian and Poisson channels, respectively.

\begin{figure}
	\centering
	\vspace{-0.4cm}
	\begin{subfigure}[b]{\figWidth}
		\centering
		\includegraphics[width=\figWidth]{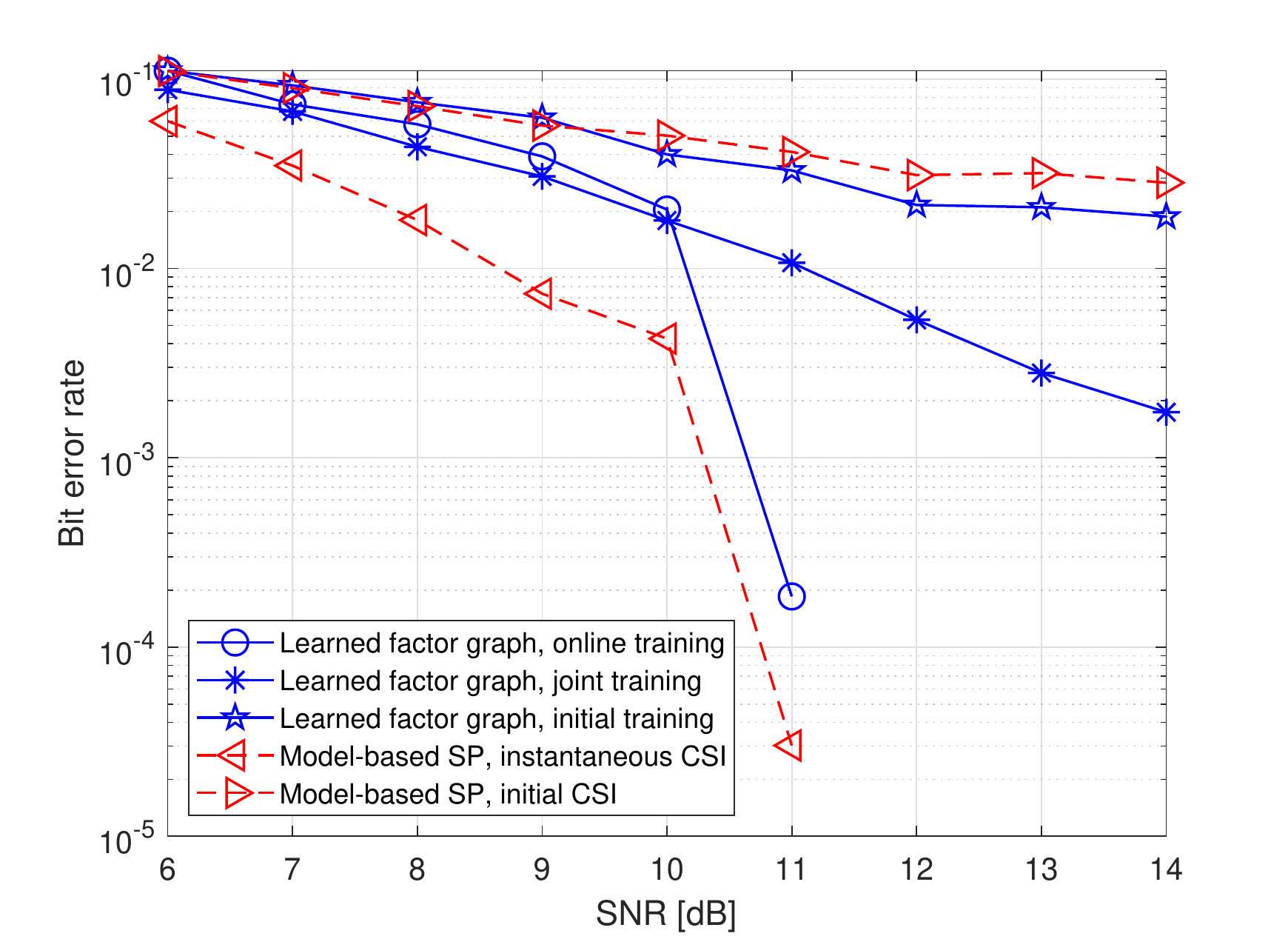}
		\caption{BER performance of different detectors in the Gaussian channel.}\label{fig:TVChannelAWGN}
	\end{subfigure}
	
	\begin{subfigure}[b]{\figWidth}
		\centering
		\includegraphics[width=\figWidth]{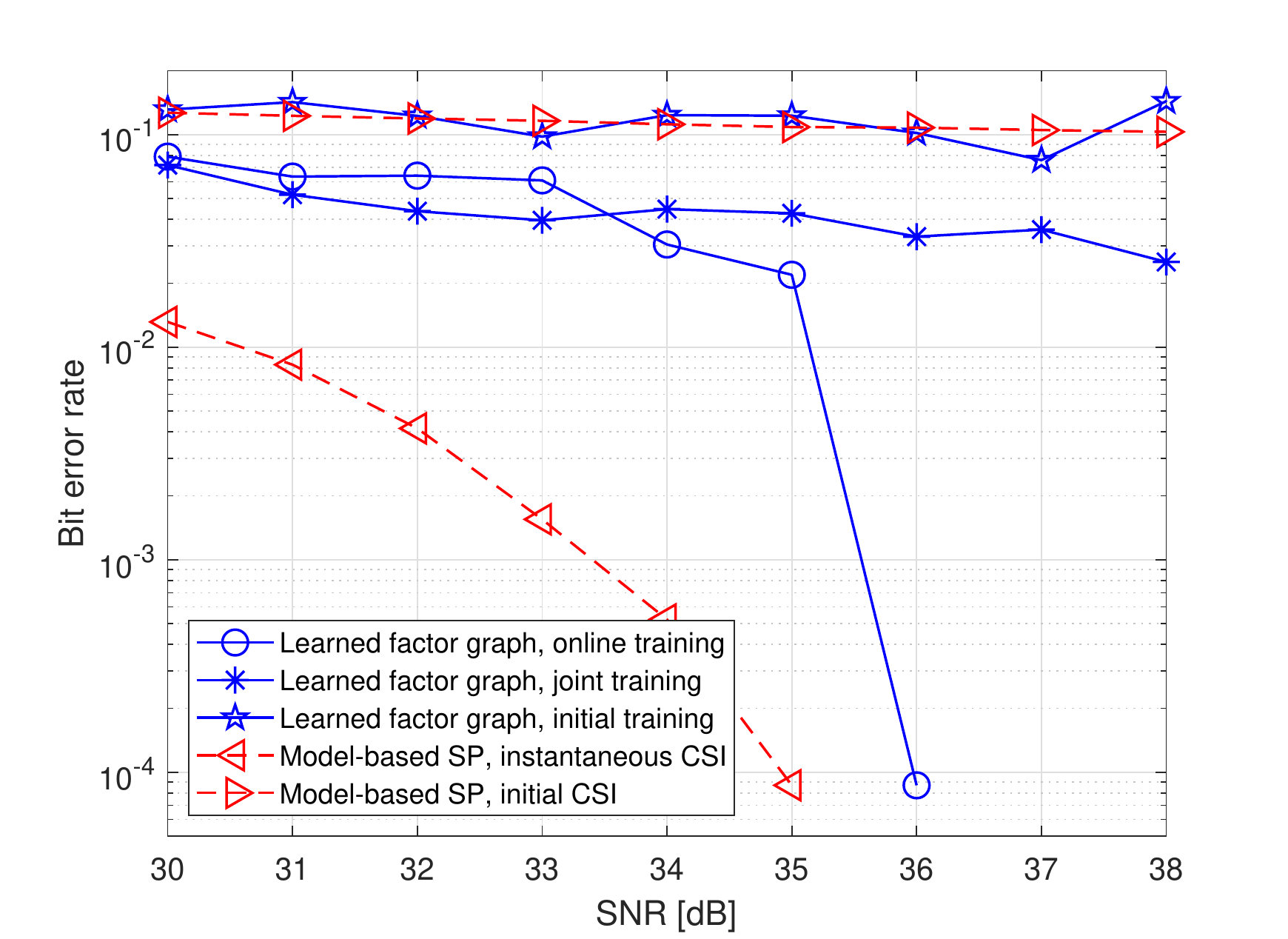}
		\caption{BER performance of different detectors in the Poisson channel.}\label{fig:TVChannelPoisson}
	\end{subfigure} 
	\caption{\ac{ber} for blockwise stationary channels.}	 
	\label{fig:TVChannel}
\end{figure}

Observing Figs. \ref{fig:TVChannelAWGN}-\ref{fig:TVChannelPoisson}, we note that for both channels, as the \ac{snr} increases, learned factor graphs with online training approaches the performance of the model-based SP detector with instantaneous \ac{csi}, which implements \ac{map} detection here. The latter require accurate knowledge of the complete input-output statistical relationship for each block. For low \acp{snr}, the performance of \ac{sp} inference over the online-trained factor graph is only slightly improved compared with training only using the initial channel. This can be explained by noting that for high \ac{snr} values, the number of symbol detection errors does not grow above the code distance as the channel changes between block,  and thus the proposed online training scheme is capable of generating reliable labels. The online-trained factor graphs can accurately track the channel, allowing inference with optimal-approaching performance. However, for low \ac{snr} values, the \ac{rs} decoder frequently fails to correctly decode the bits, and the online training methods does not frequently update its neural network, thus achieving  only a minor improvement over using only the initial  training data set. The ability to track channel variations in a decision-directed manner can be potentially improved by utilizing syndrome decoders, as proposed in \cite{teng2020syndrome}. We leave this study for future work.

We also observe in Figs. \ref{fig:TVChannelAWGN}-\ref{fig:TVChannelPoisson} that the (offline) joint training approach allows inference over learned factor graphs to achieve improved \ac{ber} performance compared with using only initial training. This follows since the resulting decoder is capable of operating in a broader range of different channel conditions. Still, joint training is notably outperformed by the \ac{sp} detector with instantaneous \ac{csi}, whose \ac{ber} performance is approached only when using online training at high \acp{snr}. 

The results reported in Figs. \ref{fig:TVChannelAWGN}-\ref{fig:TVChannelPoisson} indicate that in order to reliably cope with non-stationary conditions, learned factor graphs designed for stationary setups should be combined with additional mechanisms for tracking the statistical variations, such as the online training mechanism discussed in Subsection~\ref{subsec:OnlineTrain}. Nonetheless, one can also design learned factor graphs to cope with non-stationarity by learning different function nodes for different time instances, at the cost of requiring larger datasets for training and possible limitations on the duration of the time sequences. We leave this extension of learned factor graphs for future study.

\vspace{-0.2cm}
\subsection{Inference Computational Complexity}
\label{subsec:CompComplex}
\vspace{-0.1cm}
We conclude this section by comparing the computational complexity of inference over learned factor graphs with that of end-to-end deep learning models. Specifically, we focus on the symbol detection setup detailed in Subsection~\ref{subsec:symdet} and consider the \ac{sbrnn} algorithm as the end-to-end deep learning model, as this architecture is shown to achieve the best performance among the end-to-end deep learning models evaluated in Figs.~\ref{fig:AWGN}-\ref{fig:Poisson}.  

Recall that $\Blklen$ is the length of the sequence, $\mathcal{S}$ is the symbol set, and $l$ is the memory length of the channel. Let $N_{f}$ be the computational complexity of the learned factor node in the factor graph, i.e., the number of operations required to map a single observation $y_i$ and state vector $\myVec{s}_{i-1}$ into an estimate of the function node $\{\hat{f}(y_i, \myVec{s}_i,\myVec{s}_{i-1})\}_{s_i \in \mySet{S}}$. For instance, using the classification \ac{dnn} architecture illustrated in Fig.~\ref{fig:LearnedFunctionNode}(a), this operation involves passing the value $y_i$ through a fully-connected \ac{dnn}, which using the \ac{dnn} in Fig.~\ref{fig:3FC_Architecture} is comprised of approximately $6\cdot 10^3$ multiplications, and computing $|\mySet{S}|$ transition estimates from $\myVec{s}_i$ using a look-up table. Once the messages are computed for each state and for each time instance $i$, i.e., $|\mySet{S}|^{l}\cdot \Blklen$  times, inference is carried out by computing the forward and backward messages via \eqref{eqn:Recursion1Forwards}-\eqref{eqn:Recursion1Backwards}, resulting in an overall computational burden of the order of  $\mathcal{O}(N_f \cdot |\mathcal{S}|^{l}\cdot  t)$. For comparison, letting $N_{r}$ be the computational complexity of the \ac{rnn} block corresponding to one time-step in the \ac{sbrnn} architecture, the computational complexity of the \ac{sbrnn} is on the order of $\mathcal{O}(N_{r} \cdot l\cdot  t)$. Generally, we expect $N_{r} \gg N_{f}$; For example, in the scenario reported in Subsection~\ref{subsec:symdet} computing the function nodes is done using $N_f\approx 6\cdot 10^3$ multiplications, while the \ac{sbrnn} architecture requires $N_r \approx 1.3\cdot 10^5$ multiplications per time instance. Therefore, for the symbol detection scenario, \ac{sp} over learned factor graphs does not only train with less data compared with the \ac{sbrnn} system, but also infers at reduced computational complexity.

When the state cardinality $|\mathcal{S}|^{l+1}$ is large, \ac{sp} inference, and thus also its application over learned factor graphs, may be computationally prohibitive. In these regimes, one can still utilize learned factor graphs for inference at controllable computational burden by utilizing alternative inference methods based on, e.g., state reduction  \cite{lin04VDbeam}. This indicates that it is possible to have learned factor graphs that are efficient both in terms of sample efficiency during training, and in terms of computational complexity during inference. 
Nonetheless, we leave the detailed study of the combination of learned factor graphs with such reduced complexity inference methods for future work. 

\color{black} 

\vspace{-0.2cm}
\section{Conclusions}
\label{sec:Conclusions}
\vspace{-0.1cm}
In this work we proposed a framework for inference from stationary time sequences via learned factor graphs, combining the model-based \ac{sp} algorithm with data-driven \ac{ml} tools. By exploiting domain knowledge of a stationary and Markovian characteristics, encountered in many applications in signal processing and communications, the factor graph encapsulating the underlying distribution can be learned separately from the overall inference task. This results in a hybrid model-based/data-driven system based on compact neural networks which can be trained with relatively small training sets. The resulting algorithm carries out inference over the learned factor graph in a manner which is not restricted to a specific number of input samples. The integration of \acp{dnn} for learning the function nodes combined with domain knowledge which determines the structure of the graph results in a system which learns from data to carry out \ac{map}-approaching detection in complex setups. Furthermore, a learned factor graph can be used with different message passing based inference algorithms, other than the \ac{sp} method.   Our numerical evaluations demonstrate the ability of learned factor graphs to facilitate accurate inference and improve upon existing classifiers for sleep pattern detection and symbol recovery. We also show that its ability to train with small training sets enables personalized learning as well as tracking of blockwise temporal variations in the statistical model, which are both extremely challenging to carry out using conventional highly-parameterized \acp{dnn}. 

\bibliographystyle{IEEEtran}
\bibliography{IEEEabrv,refs}
	
\end{document}